\documentclass[10pt,twocolumn,letterpaper]{article}

\usepackage{cvpr}              %

\usepackage{graphicx}
\usepackage{amsmath}
\usepackage{amssymb}
\usepackage{booktabs}
\usepackage{xcolor}

\usepackage{amsmath,amsfonts,bm}

\def\eqref#1{equation~\ref{#1}}

\def\1{\bm{1}}

\def\rvc{{\mathbf{c}}}
\def\rvd{{\mathbf{d}}}

\def\rvo{{\mathbf{o}}}

\def\rvr{{\mathbf{r}}}

\def\rvx{{\mathbf{x}}}

\def\rvz{{\mathbf{z}}}

\def\rmC{{\mathbf{C}}}

\def\rmI{{\mathbf{I}}}

\def\rmR{{\mathbf{R}}}

\DeclareMathAlphabet{\mathsfit}{\encodingdefault}{\sfdefault}{m}{sl}
\SetMathAlphabet{\mathsfit}{bold}{\encodingdefault}{\sfdefault}{bx}{n}

\usepackage[pagebackref,breaklinks,colorlinks]{hyperref}
\usepackage{appendix}

\usepackage[capitalize]{cleveref}
\crefname{section}{Sec.}{Secs.}
\Crefname{section}{Section}{Sections}
\Crefname{table}{Table}{Tables}
\crefname{table}{Tab.}{Tabs.}

\def\confName{CVPR}
\def\confYear{2022}

\newcommand{\AT}[1]{\textcolor{black}{#1}}

\begin{document}

\title{Disentangled3D: Learning a 3D Generative Model with Disentangled Geometry and Appearance from Monocular Images}

\author{Ayush Tewari$^{1,2}$~~~Mallikarjun B R$^{1}$~~~Xingang Pan$^{1}$~~~Ohad Fried$^{3}$ \\ \vspace{0.4cm}
Maneesh Agrawala$^{4}$~~~Christian Theobalt$^{1}$\\ 
		$^1$Max Planck Insitute for Informatics~~~$^2$MIT~~~$^3$Interdisciplinary Center, Herzliya~~~$^4$Stanford University
}
\twocolumn[{%
\renewcommand\twocolumn[1][]{#1}%
\maketitle
\begin{center}
\includegraphics[width=0.96\linewidth]{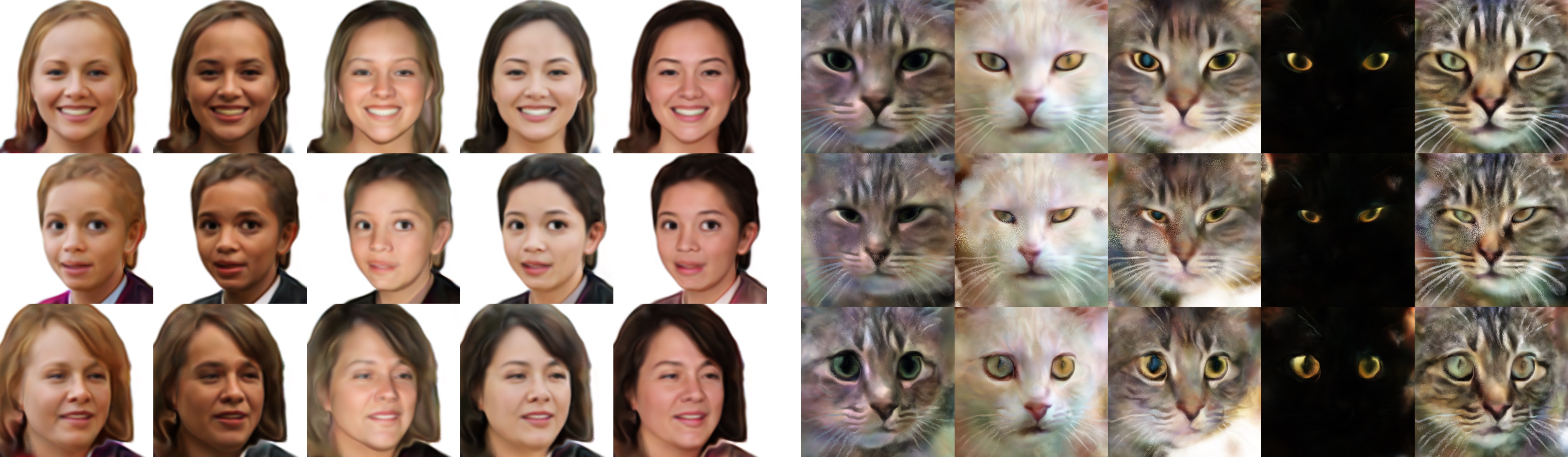}
\captionof{figure}{
    Our model can disentangle geometry, appearance, and pose in synthesized images. 
    This figure visualizes results on FFHQ~\cite{Karras_2019_CVPR} (first 5 columns) and Cats~\cite{zhang2008cat} (last 5 columns).
    Each row shows images rendered with the same pose and geometry, but with different appearances. 
Each column shows images rendered with different poses and geometry, but with the same appearance. 
    }
    \label{fig:teaser}
	\end{center}
}]

\begin{abstract}

Learning 3D generative models from a dataset of monocular images enables self-supervised 3D reasoning and controllable synthesis. 
State-of-the-art 3D generative models are GANs that use neural 3D volumetric representations for synthesis. 
Images are synthesized by rendering the volumes from a given camera. 
These models can disentangle the 3D scene from the camera viewpoint in any generated image. 
However, most models do not disentangle other factors of image formation, such as geometry and appearance. 
In this paper, we design a 3D GAN which can learn a disentangled model of objects, just from monocular observations. 
Our model can disentangle the geometry and appearance variations in the scene, i.e., we can independently sample from the geometry and appearance spaces of the generative model. %
This is achieved using a novel non-rigid deformable scene  formulation. 
A 3D volume that represents an object instance is computed as a non-rigidly deformed canonical 3D volume. 
Our method learns the canonical volume, as well as its deformations, jointly during training. 
This formulation also helps us improve the disentanglement between the 3D scene and the camera viewpoints using a novel pose regularization loss defined on the 3D deformation field. 
In addition, we  model the inverse deformations, enabling the computation of dense correspondences between images generated by our model. 
Finally, we design an approach to embed real images into the latent space of our model, enabling  editing of real images. 

\end{abstract}
\section{Introduction}

State-of-the-art generative models directly operate in the  image space using 2D CNNs. 
These models, such as StyleGAN and its variants~\cite{Karras_2019_CVPR,karras2020analyzing,Karras2021} have achieved a high level of photorealism.
However, image-based models do not offer direct control over the underlying 3D scene parameters, such as camera and geometry. 
While some methods add camera viewpoint control over pretrained image-based GAN models~\cite{deng2020disentangled,tewari2020stylerig,FreeStyleGAN2021,mallikarjun2021photoapp}, the results are limited by the quality of 3D consistency  of the pretrained models.  %

In contrast to the image-based methods, recent approaches  learn GAN models directly in the 3D space~\cite{Schwarz2020NEURIPS,niemeyer2021giraffe,chanmonteiro2020pi-GAN,gu2021stylenerf,nguyen2019hologan}. 
In this case, the generator network synthesizes a 3D representation of the scene as output, which can then be rendered from a virtual camera to generate the image.
Since the 3D scene is explicitly modeled, the camera parameters  are disentangled from the scene itself in the image synthesis process. 
However, other scene properties such as geometry and appearance remain entangled and cannot be  controlled independently. 
While some 3D GAN approaches have attempted to disentangle geometry from appearance~\cite{Schwarz2020NEURIPS,niemeyer2021giraffe}, their design choices are not physically-motivated, which leads to inaccurate solutions where appearance information can leak through the geometry component. 
In contrast, our proposed approach is inspired by recent non-rigid formulations for novel viewpoint  synthesis of dynamic scenes~\cite{tretschk2021nonrigid,park2021nerfies}. 
These methods model the deformations in a scene observed across time, by separating the 3D reconstruction of each frame into a canonical 3D reconstruction and its deformations.  
Yet, even though these methods can learn to synthesize novel viewpoints of a deforming scene, they are limited to modeling a single scene, and they cannot control the appearance of the scene. 

In this work, we propose D3D, a GAN with two separate and independent components for geometry and appearance. 
We extend the non-rigid formulation to the case of modeling multiple instances of a deformable object category, such as human heads, cats, or cars. %
Each instance of the object class is modeled as a deformation of a canonical volume, which is shared across the object category. %
Our method learns the canonical volume, as well as the instance-specific geometric deformations jointly from datasets of monocular images. 
The canonical volume has a fixed geometry while its appearance can be changed independent of the geometric deformations.
{This formulation by design motivates disentanglement between the geometric deformations and appearance variations, which has been a challenging task, especially as we are limited to monocular images for training. }

In addition to the disentanglement of geometry and appearance, our formulation allows for other advantages over state-of-the-art methods. 
Since our geometric deformations are explicit Euclidean transformations, we can enforce useful properties in the model, such as pose consistency over the generated 3D volumes. 
Existing 3D GANs do not always manage to disentangle the camera viewpoint and the generated 3D volumes, especially when the hand-crafted prior camera distribution does not match the real distribution of the training dataset. %
We design a pose regularization loss, which can enforce the consistency of the object pose, improving the quality of camera and scene disentanglement. 
In addition, we learn an inverse deformation network, allowing us to compute dense correspondences between images generated by our model. 
Finally, we allow editing of input photographs using D3D by  mapping a given image to the corresponding geometry and appearance latent codes, as well as the camera pose.
In summary, this paper presents the following contributions:
\begin{enumerate}
\setlength\itemsep{0em}
    \item A generative model which can disentangle geometry,  appearance, and camera pose in the generated images. 
    This is enabled by a generalization of the non-rigid scene formulation to deformable object categories. 
    \item A novel training framework for 3D GANs, which enables  pose consistency of the generated volumes, as well as the computation of dense correspondences between generated images. 
    \item Editing of real images by computing their embedding in our GAN space. This enables intuitive control over the camera pose, appearance and geometry in images.
\end{enumerate}
\section{Related Work}

\subsection{3D Generative Adversarial Networks}
2D Generative adversarial networks (GANs)~\cite{goodfellow2014generative} have achieved great success in synthesizing high-fidelity images, but lack explicit control over scene parameters, and do not guarantee 3D consistency.
Several attempts have been made to incorporate GANs with 3D representations for 3D-aware image synthesis.
Some works directly train on 3D data~\cite{wu2016learning,chen2021decor}, while others only use 2D images by leveraging differentiable 3D-2D projection~\cite{nguyen2019hologan,nguyen2020blockgan,liao2020towards,niemeyer2021giraffe,henzler2019escaping,szabo2019unsupervised,Schwarz2020NEURIPS,chanmonteiro2020pi-GAN,hao2021GANcraft}.
In this work, we focus on the latter paradigm, which is more practical, as collecting 3D scans is resource-intensive. %
Many methods~\cite{nguyen2019hologan,nguyen2020blockgan,liao2020towards,niemeyer2021giraffe,hao2021GANcraft} synthesize 3D features which are converted into the final images using image-based networks. 
This limits the quality of 3D consistency in the rendered results.
Henzler et al.\,\cite{henzler2019escaping} and Szabo et al.\,\cite{szabo2019unsupervised} learn to generate explicit 3D voxels and meshes respectively, but produce shapes and images with limited quality.
Recently, there has been a surge of interest in adopting coordinate-based neural volumetric representations~\cite{mildenhall2020nerf}, defined using MLPs, as the 3D representation for GANs~\cite{Schwarz2020NEURIPS,chanmonteiro2020pi-GAN,pan2021shadegan,xu2021generative}.
These approaches have achieved high-quality 3D-aware image synthesis with high-quality 3D consistency.
However, the disentanglement between geometry and appearance has not been fully explored.

\subsection{Disentanglement}

\paragraph{Monocular Approaches:}
Zhu et al.~\cite{zhu2018visual} proposed a GAN that can disentangle the shape, appearance, and camera variations in images.
The final appearance is synthesized using a 2D network, which can limit the 3D consistency in the synthesized images. 
The closest approach to our work is GRAF~\cite{Schwarz2020NEURIPS}. 
The network consists of a shared backbone MLP, with separate color and density heads. 
The appearance latent code is provided as an input to the color head, while the shape latent code is provided as an input to the backbone. 
The backbone MLP corresponds to the deformation network in our design. 
However, unlike our deformation network, GRAF does not explicitly model 3D deformations, and the output of the backbone network lives in a higher-dimensional space. 
This leads to lower-quality disentanglement, where the color information can leak into the backbone network, and the appearance code can be ignored. 
Unlike GRAF, our framework also enables the computation of dense correspondences, which is made possible by our explicit modeling of the forward and inverse deformation fields.
GIRAFFE~\cite{niemeyer2021giraffe} uses the same disentanglement strategy as GRAF, however, it also relies on a 2D rendering network which limits 3D consistency. 

\paragraph{Multi-View: }
Other approaches disentangle these factors using multi-view imagery. 
Multi-view images provide more information about the 3D geometry which makes this task easier. 
Xiang~\etal~\cite{xiang2021neutex} proposed NeuTex, which can disentangle the shape from appearance by learning the appearance information on a texture map.
The mapping between the 3D scene coordinates and 2D texture coordinates is also learned by the method. 
However, NeuTex is scene-specific and is thus not a generative model, i.e., we cannot randomly sample realistic scenes from their model. 
Liu~\etal~\cite{liu2021editing} proposed a method for editing radiance fields. 
Their network is trained on a class of objects and enables controllable editing at test time. 
CodeNeRF~\cite{jang2021codenerf} also achieves independent control over the shape and appearance components. 
Both these approaches share a similar design choice with GRAF, i.e., their canonical shape space does not receive a 3D input. Instead, it lives in a higher-dimensional space, which is not interpretable. 
Our method, in contrast, is physically inspired, as it models explicit 3D deformations between different object instances.
In addition, our method is the only one that enables dense correspondences between synthesized images. %

\subsection{Non-Rigid NeRFs}

Another category of papers~\cite{xian2021space,pumarola2020d,park2021nerfies,tretschk2021nonrigid,li2021neural} addresses the problem of time-varying novel-view synthesis given monocular videos.
Xian~\etal~\cite{xian2021space} extend the NeRF formulation to parameterize the network with time to model time-dependent view interpolation.
D-NeRF~\cite{pumarola2020d}, NR-NeRF~\cite{tretschk2021nonrigid}, and Nerfies~\cite{park2021nerfies} learn a canonical representation of the entire scene from which the other frames can be obtained by  learning deformations to the canonical space.
These methods also propose a number of regularizers to control the deformation space.
Li~\etal~\cite{li2021neural} takes a different approach by learning a 3D flow field between neighbouring time samples. They supervise their method with 2D optical flow and depth predictors. 
In contrast to these approaches, our method is a generative model and is not limited to a given scene. In addition, we can also disentangle appearance from geometry.

\section{Method}
\begin{figure*}
\centering
\includegraphics[width=1.0\textwidth]{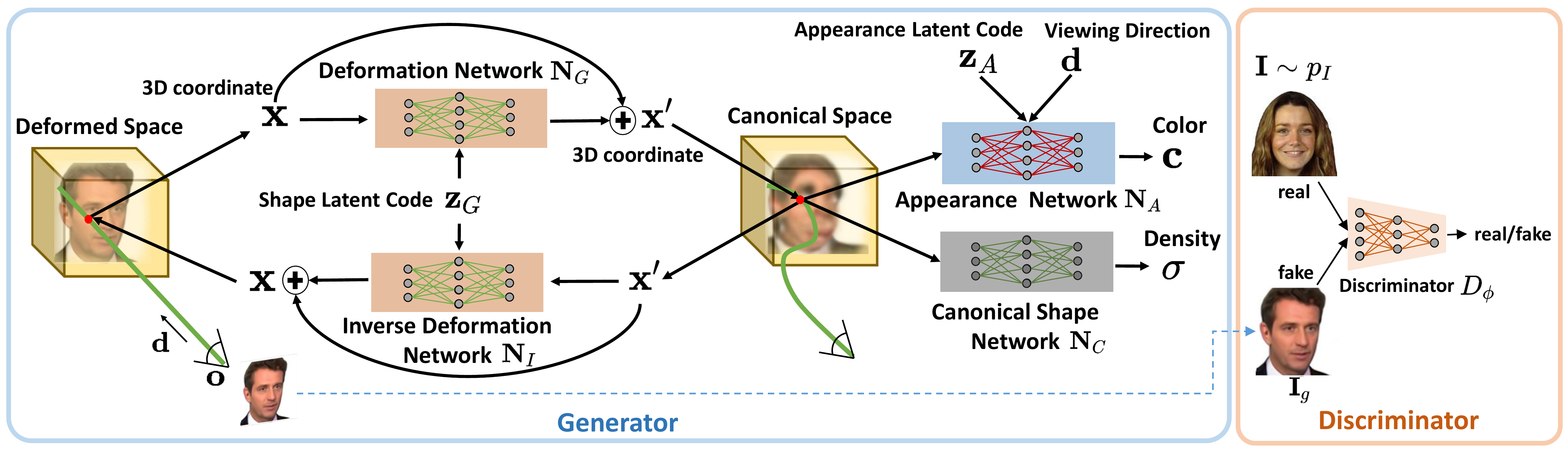}
\vspace{-0.5cm}
\caption{\textbf{Method overview.} Our generator consists of three main components: 1) a deformation network $\mathbf{N}_G$ that maps the coordinates from deformed space to the canonical space conditioned on a shape latent code $\mathbf{z}_G$, 2) a canonical shape network $\mathbf{N}_C$ that models the canonical volume density, and 3) an appearance network $\mathbf{N}_A$ that models the color of the canonical space conditioned on a color latent code $\mathbf{z}_A$. We can optionally incorporate a inverse deformation network $\mathbf{N}_I$ that models the inverse deformation so that dense correspondence could be obtained. Images are generated by performing volume rendering in the deformed space. 
A discriminator $D_\phi$ is used for adversarial training.
The terms color and appearance are used interchangeably in the paper. }
\vspace{-0.1cm}
\label{fig:pipeline}
\end{figure*}

We use a neural volumetric representation to represent objects, i.e., an MLP network encodes the 3D coordinates and regresses the density and radiance values of the 3D volume~\cite{mildenhall2020nerf}. 
The output volume can be rendered from a virtual camera using volumetric integration to produce the final image. 
The network is trained in an adversarial manner using monocular images as the training data. 

\subsection{Network Architecture}
\label{sec:3.1}
The pipeline of our method is shown in Fig.~\ref{fig:pipeline}, which includes a generator and a discriminator. 
Since we want to disentangle the geometry and appearance in the scene, we model these components as individual MLP networks, represented as functions $\mathbf{N}_G(\cdot)$ and $\mathbf{N}_A(\cdot)$ . %
In addition, we use another MLP network, represented as function $\mathbf{N}_C(\cdot)$,  to model the canonical object shape. 
{For any object class, a shared canonical volume defined by $\mathbf{N}_C(\cdot)$ will represent a canonical geometry. 
$\mathbf{N}_G(\cdot)$ will model the deformation of a specific object instance with respect to the canonical geometry, and $\mathbf{N}_A(\cdot)$ will represent the color of the canonical volume.}
Furthermore, we can optionally train an inverse deformation network $\mathbf{N}_I(\cdot)$ that models the inverse mapping of $\mathbf{N}_G(\cdot)$, enabling  dense correspondence (introduced in Sec.~\ref{sec:inverse}).
Next, we introduce these components in detail.

Our method models color and volume density in the 3D space.
For a point with coordinate $\mathbf{x} \in \mathbb{R}^3$, we first send it to the deformation network $\mathbf{N}_G(\cdot)$ to obtain its corresponding point $\mathbf{x'} \in \mathbb{R}^3$ in the canonical space as
\begin{align}
    \mathbf{x'}(\mathbf{x}, \mathbf{z}_G) = \mathbf{N}_G(\mathbf{x}, \mathbf{z}_G) + \mathbf{x} \,, \label{eq:deformation}
\end{align}
where $\mathbf{z}_G \in \mathbb{R}^{256}$ is the geometry latent vector sampled from a Gaussian distribution.
Thus, $\mathbf{z}_G$ represents different object shapes by varying the deformation field.
We can compute the volume density $\sigma \in \mathbb{R}^+$ in the canonical space as:
\begin{align}
    \sigma(\mathbf{x}, \mathbf{z}_G) = \mathbf{N}_C(\mathbf{x'}(\mathbf{x}, \mathbf{z}_G)) \,.
    \label{eq:density}
\end{align}
where the canonical network $\mathbf{N}_C$ does not receive any conditioning other than the input coordinate. 

Next, we represent the view-dependent color, i.e., radiance, of the scene in the canonical space as:
\begin{align}
    \mathbf{c}(\mathbf{x}, \rvd, \mathbf{z}_G, \mathbf{z}_A) = \mathbf{N}_A(\mathbf{x'}(\mathbf{x}, \mathbf{z}_G), \rvd, \mathbf{z}_A) \,.
    \label{eq:color}
\end{align}
Here, $\mathbf{c}(\mathbf{x}, \rvd, \mathbf{z}_G, \mathbf{z}_A) \in \mathbb{R}^3$, $\rvd \in \mathbb{S}{^2}$ is the viewing direction, and $\mathbf{z}_A$ is a randomly sampled $256$ dimensional vector. %
Thus, we can vary the color without changing geometry by simply sampling different color latent vectors $\mathbf{z}_A$.

\paragraph{Disentanglement}
The explicit modeling of deformation fields in our model by design encourages the disentanglement between the geometry and appearance components.
Specifically, our geometry deformation network generates 3-dimensional Euclidean transformations, which is added to the input coordinate $\rvx$ to obtain the deformed coordinate in the canonical space.
This is in contrast with the state-of-the-art methods~\cite{Schwarz2020NEURIPS,niemeyer2021giraffe}, 
which use a similar network architecture, but their backbone network directly produces a high-dimensional output %
without any physical interpretation. 
This design choice hinders good disentanglement, as this high-dimensional space can also encode information about the color of the object. 
In contrast, our formulation strictly restricts the output of the geometry network to a 3-dimensional vector that models a  coordinate offset.
This makes it less likely for our method to leak color information compared to previous methods.

While our formulation discourages the color information from leaking into the geometry channel, %
this approach does not completely resolve all geometry-appearance ambiguities. 
Consider the domain of human heads where the distinct states of mouth open and mouth closed can be represented in two ways: one where the geometry component is responsible for this deformation, another, where the geometry stays the same, and the color component changes instead.
While only the first solution is physically correct, both geometry and appearance changes can plausibly lead to realistic images. 
Note that we do not have 3D information to judge the physically correct 3D solution---we only rely on monocular images. 
This ambiguity cannot be resolved solely by the separation of geometry and appearance channels into separate networks.
Thus, we additionally control the level of disentanglement by using different sizes of networks for the geometry and appearance components. 
Specifically, when the appearance network is too large, face expression changes like mouth open would tend to be represented by the appearance network as it is easier to optimize.
Balancing the depths of the deformation and appearance networks ensures good disentanglement for all datasets. 

\subsection{Volumetric Integration}
We use the volumetric neural rendering formulation, following NeRF~\cite{mildenhall2020nerf}.
Unlike NeRF that has multiple views of the same scene and their corresponding poses, we only have unposed monocular images.
Thus, during training, a virtual camera pose is first sampled from a prior distribution.
To render an image under a given camera pose, each pixel color $\rmC$ is computed via volume integration along its corresponding camera ray $\rvr(t) = \rvo + t\rvd$ with near and far bounds $t_n$ and $t_f$ as below:
\begin{align}
\rmC(\rvr) &= \int_{t_n}^{t_f} T(t) \sigma(\rvr(t)) \rvc(\rvr(t)) dt \nonumber \\
\text{where} \quad & T(t) = \text{exp}(-\int_{t_n}^{t}\sigma(\rvr(s)) ds).
\label{eq:nerf}
\end{align}
Here the dependence of $\sigma$ and $\rvc$ on $\rvz_G$ and $\rvz_A$ is omitted for clarity.
In practice, we implement a discretized numerical integration using stratified and hierarchical sampling, following NeRF~\cite{mildenhall2020nerf}.
For each sampled discrete point along the ray, we obtain $\sigma$ and $\rvc$ by querying our generator according to Eq.(\ref{eq:density}) and Eq.(\ref{eq:color}).
With this volumetric rendering, we can render an image $\rmI_g$ under any camera pose $\bm{\xi}$ using our model.
We summarize this process as $\rmI_g = G_{\theta} (\rvz_G, \rvz_A, \bm{\xi})$, where the generator $G_{\theta}$ includes the  $\mathbf{N}_G$, $\mathbf{N}_A$, and $\mathbf{N}_C$ components  mentioned earlier, and $\theta$ denotes the learnable parameters.
This rendering process is differentiable and thus can be trained using backpropagation.

\subsection{Loss Functions}
\paragraph{Adversarial Loss}
We train our generator $G_{\theta}$ along with a discriminator $D_\phi$ with parameters $\phi$ using an adversarial loss.
We use the discriminator architecture from $\pi$-GAN~\cite{chanmonteiro2020pi-GAN}.
During training, the geometry latent vector $\rvz_G$, color latent vector $\rvz_A$, and camera pose $\bm{\xi}$ are randomly sampled from their corresponding prior distributions to generate fake images, while real images $\rmI$ are sampled from the training dataset of distribution $p_\mathcal{D}$.
Our model is trained with a non-saturating GAN loss~\cite{mescheder2018training} as:
\begin{align}
\mathcal{L_\text{adv}}(\theta, \phi) &= f \Big( D_{\phi} (G_{\theta} (\rvz_G, \rvz_A, \bm{\xi})) \Big)  \nonumber \\
&+ f(-D_{\phi}(\rmI)) + \lambda \|\nabla D_{\phi}(\rmI)\|^2,
\label{eq:gan}
\end{align}
where $f(u) = -\log(1 + \exp(-u))$, and $\lambda$ is the coefficient for $R_1$ regularization.
In practice, $\rvz_G$, $\rvz_A$, $\bm{\xi}$, and $\rmI$ are randomly sampled as mini-batches, which is an approximation of taking expectation over these variables.

\paragraph{Pose Regularization}
With the adversarial loss, the generator learns to synthesize realistic images, when rendered from camera poses sampled from the manually specified prior camera distribution. %
Ideally, the network learns to disentangle the pose and the 3D scene in the generated images, i.e., the generated volumes are in a consistent pose. 
However, in many cases, the network converges to a solution where the generated volumes have the objects in different poses. 
This is usually the case when the prior distribution over camera poses is inaccurate. 

In our formulation, the explicit modeling of the deformation field makes it possible to enforce pose consistency of the generated volumes. 
To achieve this, we first compute the global rotation component $R \in SO(3)$ of the deformation field $\mathbf{D}(\mathbf{x}, \mathbf{z}_G)$ using SVD %
orthogonalization~\cite{levinson2020analysis}. %
Here we only consider sampled points $\mathbf{x}$ with a rendering weight (the scalar factor applied to the color of a 3D point during integration) greater than a specified threshold. %
Our pose regularization loss term is then computed as 
\begin{align}
    \mathcal{L_\text{pose}}(\theta) = \| R - I \|^2 ,
\end{align}
where $I$ is the identity matrix.  %
We use a differentiable SVD implementation which allows training using backpropagation. 
This term is very different from the regularization terms introduced in existing non-rigid formulations~\cite{tretschk2021nonrigid,park2021nerfies}, where local deformations are encouraged to be rotations. 
This is not suitable in our case, as we are modeling deformations across object instances, which can include stretching, compression, and discontinuities. 
Our loss term, on the other hand, encourages the deformations to not include any global rotation, which gives rise to a disentangled solution where the camera pose variation accounts for all pose changes in the rendered images. 

We first train our networks with a combination of the two loss functions 
\begin{align}
    \label{eq:firststage}
    \mathcal{L}(\theta, \phi) = \mathcal{L_\text{adv}}(\theta, \phi) + \lambda_\text{pose} \mathcal{L_\text{pose}}(\theta) .
\end{align}
Then, we further model the inverse deformation field.%

\subsection{Inverse Deformation}
\label{sec:inverse}
Our network allows us to compute dense correspondences between rendered images. 
We enable this by training an inverse deformation network $\mathbf{N}_I$ with parameters $\psi$.  %
Since we are using a volumetric representation, multiple points in the volume are responsible for the color at any pixel. 
Dense correspondences, where a pixel in an image has a correspondence with only one pixel in another image, is not trivial to define. 
Thus, we simplify the formulation for the training of the inverse network by limiting its domain to points around the expected surface of the volume, which can be obtained by taking the expectation of depth using the volume rendering weights.
For any such point $\mathbf{x}$, we can compute the canonical coordinate $\mathbf{x'}(\mathbf{x}, \mathbf{z}_G)$ via Eq.~\ref{eq:deformation} and use the inverse network to go back to the deformed space as $\mathbf{x}_I = \mathbf{N}_I (\mathbf{x'}(\mathbf{x}, \mathbf{z}_G), \mathbf{z}_G) + \mathbf{x'}(\mathbf{x}, \mathbf{z}_G)$. 
We can formulate the following constraint on the inverse deformation network:
\begin{align}
    \label{eq:inverse}
    \mathcal{L_\text{inv}}(\psi) = \| \mathbf{x}_I - \mathbf{x} \|^2 + \lambda_\text{img} \| \rmR(\mathbf{x}_I) - \rmR(\mathbf{x}) \|^2 .
\end{align}
Here, $\rmR$ is a rendered image of the volume at the resolution being used for training.  
$\mathbf{x}$ are sampled from the image using the expected depth value. 
$\rmR(\mathbf{x})$ is an operation that computes the color at the pixel which $\mathbf{x}$ projects to, using bilinear interpolation. 
The first term in Eq.~\ref{eq:inverse} penalizes 3D geometric deviations, while the second term can also use color information to refine the correspondences. 
After pretraining our networks with the loss as defined in Eq.~\ref{eq:firststage}, we first  train the inverse network $\mathbf{N}_I$ using $\mathcal{L_\text{inv}}$, and finally jointly train all components in our architecture with the following loss:
\begin{align}
    \label{eq:secondstage}
    \mathcal{L}(\theta, \phi, \psi) = \mathcal{L_\text{adv}}(\theta, \phi) + \lambda_\text{pose} \mathcal{L_\text{pose}}(\theta)  + \lambda_\text{inv} \mathcal{L_\text{inv}}(\psi).
\end{align}
This joint optimization of both forward and inverse deformation networks further improves dense correspondences.
Note that we do not include the inverse loss from the beginning as it can bias the deformation network to generate very small deformations, making disentanglement  challenging.

\begin{figure*}[t!]
\centering
\includegraphics[width=0.95\linewidth]{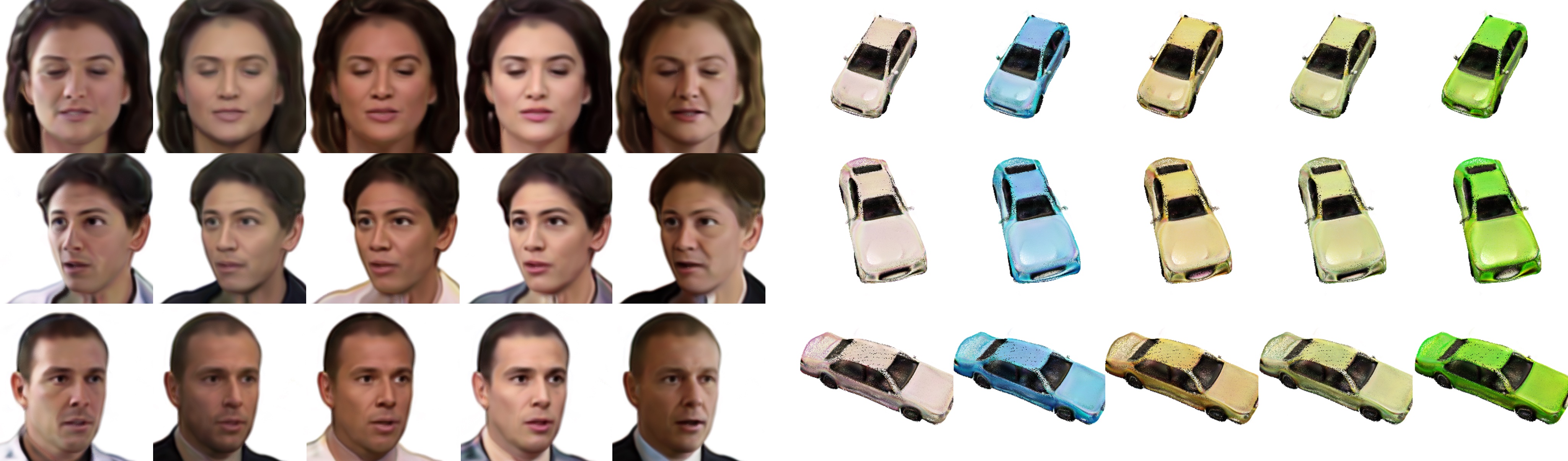}
\vspace{-0.2cm}
\caption{Qualitative results on VoxCeleb2~\cite{Chung18b}  and CARLA~\cite{dosovitskiy2017carla}. 
Each row shows images rendered with the same pose and geometry, but different appearances. 
Each column shows images rendered with different poses and geometry, but with the same appearance. 
}
\label{fig:qualitative}
\vspace{-0.2cm}
\end{figure*}

\subsection{Embedding}
Given our trained model and a real image, we could diretly optimize for the latent vector and camera pose in an iterative manner~\cite{chanmonteiro2020pi-GAN,xia2021gan}.
However, this strategy is inefficient, and can lead to lower-quality results.
We therefore learn an encoder that takes an image as input and regresses the latent vectors and camera pose.
We make use of a pre-trained ResNet~\cite{resnet_16} as our encoder backbone.
The encoder is trained on monocular images (FFHQ~\cite{Karras_2019_CVPR}), using our trained GAN as the  decoder, in a self supervised manner, using the following loss function:

\begin{align}
\label{eq:encoder}
    \mathcal{L_\text{encoder}}(\Upsilon) = \mathcal{L_\text{1}}(\Upsilon) + \lambda_\text{perc} \mathcal{L_\text{perc}}(\Upsilon) + \lambda_\text{reg} \mathcal{L_\text{reg}}(\Upsilon),
\end{align}
where, $\Upsilon$ denotes the learnable parameters of the encoder.
$\mathcal{L_\text{1}}$ is an $\ell_1$ reconstruction term, and  $\mathcal{L_\text{perc}}$ is a perceptual term defined using the features of the VGG network. 
$L_\text{reg}$ encourages the predicted  latent vectors to stay close to the average values.
The encoded results are robust, but can still miss fine-scale details.
We first  refine the results of the encoder using iterative optimization, and finally fine-tune the generator network for the given image.
We show that this strategy leads to high-quality results without degrading the disentanglement properties (see Fig.~\ref{fig:embedding}) of the generator.
Please refer to the supplemental for more details. 

\section{Results}

\paragraph{Datasets}
We demonstrate the results of our method D3D on four datasets: %
FFHQ~\cite{Karras_2019_CVPR}, VoxCeleb2~\cite{Chung18b}, Cats~\cite{zhang2008cat}, and CARLA~\cite{dosovitskiy2017carla,Schwarz2020NEURIPS}.
FFHQ and VoxCeleb2 are datasets of head portraits. %
FFHQ includes a diverse set of static images, while VoxCeleb2 is a large-scale video dataset with larger viewpoint and expression variations. 
We randomly sample a few frames from each video for VoxCeleb2. 
Cats is a dataset of cat faces, and CARLA is a dataset of synthetic cars with large viewpoint variations. 
While cars are not deformable, different car instances can be considered as deformations of a shared template. 
The instances of these datasets share a similar geometry with varying deformations, thus, they are suitable for our task. %
Since we are only interested in modeling objects, we remove the backgrounds in portrait images~\cite{yu2018bisenet}. 
However, because cat images have very little background, we do not segment them. 

\vspace{-0.3cm}
\paragraph{Training Details}
We use the same network architecture for all datasets. 
Training is done in a coarse-to-fine fashion, similar to $\pi$-GAN~\cite{chanmonteiro2020pi-GAN}. 
We use the same camera pose distribution as used in $\pi$-GAN. 
We train at $64 \times 64$ resolution on FFHQ, VoxCeleb2, and Cats, and $128 \times 128$ resolution on CARLA. 
All quantitative evaluations are performed at $128 \times 128$ resolution (once trained, images can be rendered at any resolution due to the neural scene representation). %
Please refer to the supplemental material for the hyperparameters. 
\vspace{-0.3cm}
\paragraph{Qualitative Results}

\begin{figure}
\centering
\vspace{-0.2cm}
\includegraphics[width=\linewidth]{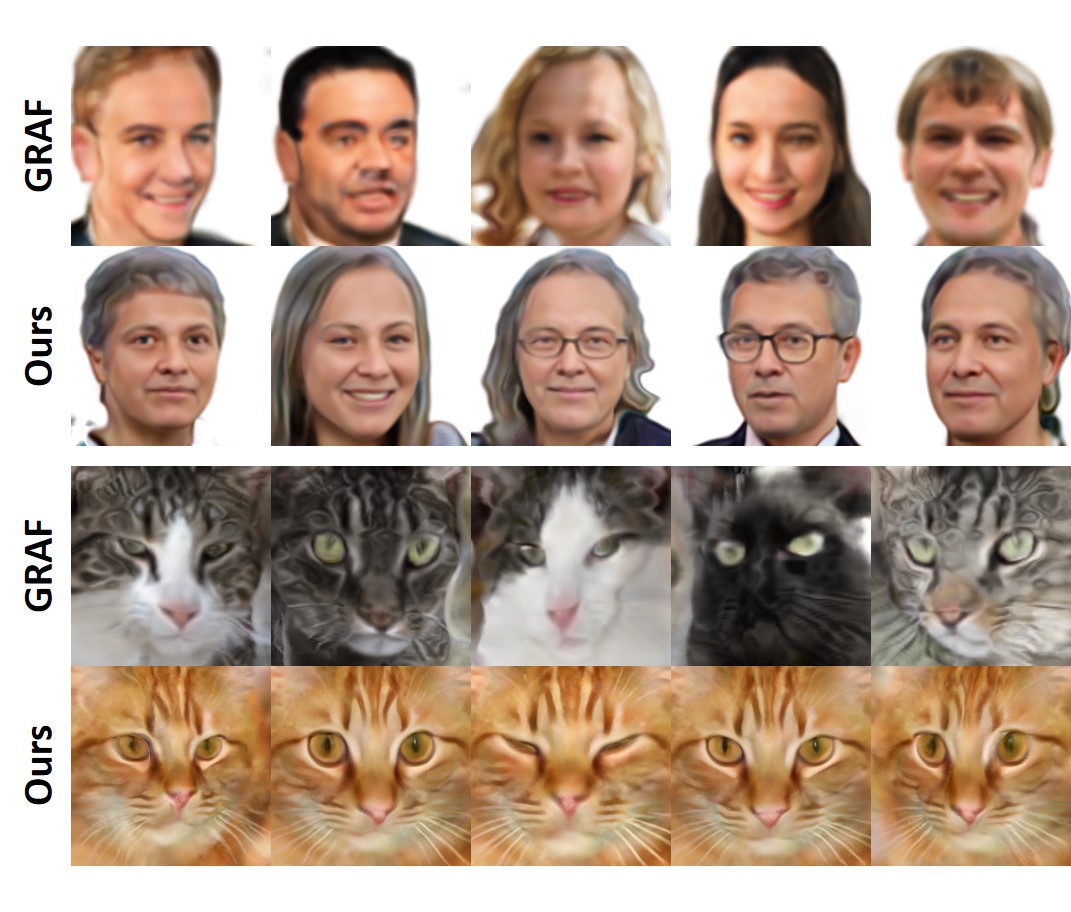}
\vspace{-0.9cm}
\caption{Comparison with GRAF on FFHQ and Cats datasets. Each row shows images rendered with a fixed appearance code and varying geometry codes. Our method can preserve the appearance better, while modeling large deformations. }
\vspace{-0.2cm}
\label{fig:graf}
\end{figure}

We first present qualitative results of our method on all four datasets in Fig.~\ref{fig:teaser} and Fig.~\ref{fig:qualitative}.  %
Our method is capable of synthesizing objects in multiple poses due to the 3D nature of the generator. 
We can disentangle the geometry and appearance variations well for all object classes. 
This is true even under challenging deformations, such as deformations due to hairstyle and mouth expressions. 
We compare the quality of disentanglement with GRAF~\cite{Schwarz2020NEURIPS} in Fig.~\ref{fig:graf}.  %
Our method significantly outperforms GRAF in terms of disentanglement. 
As explained in Sec.~\ref{sec:3.1}, GRAF also encodes appearance information in the geometry code due to the high-dimensional output of its backbone. 
In contrast, our explicit deformation enables higher-quality disentanglement. 

\begin{figure*}
\centering
\includegraphics[width=\linewidth]{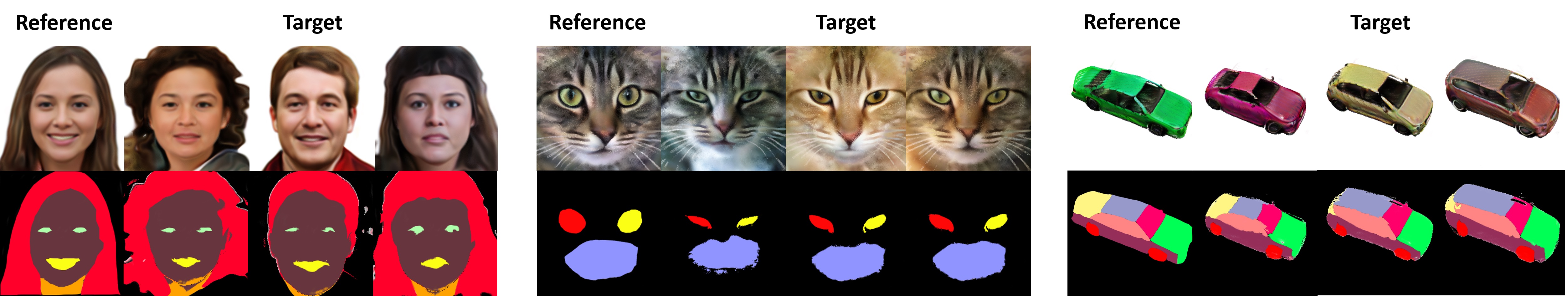}
\vspace{-0.6cm}
\caption{Our method enables dense correspondences between generated images, using the inverse deformation network. We show applications of these correspondences by transferring manual annotations on a reference image (left-most column, for each object class) to other images sampled from the model.
}
\vspace{-0.3cm}
\label{fig:correspondense}
\end{figure*}

\begin{figure}
\centering
\includegraphics[width=\linewidth]{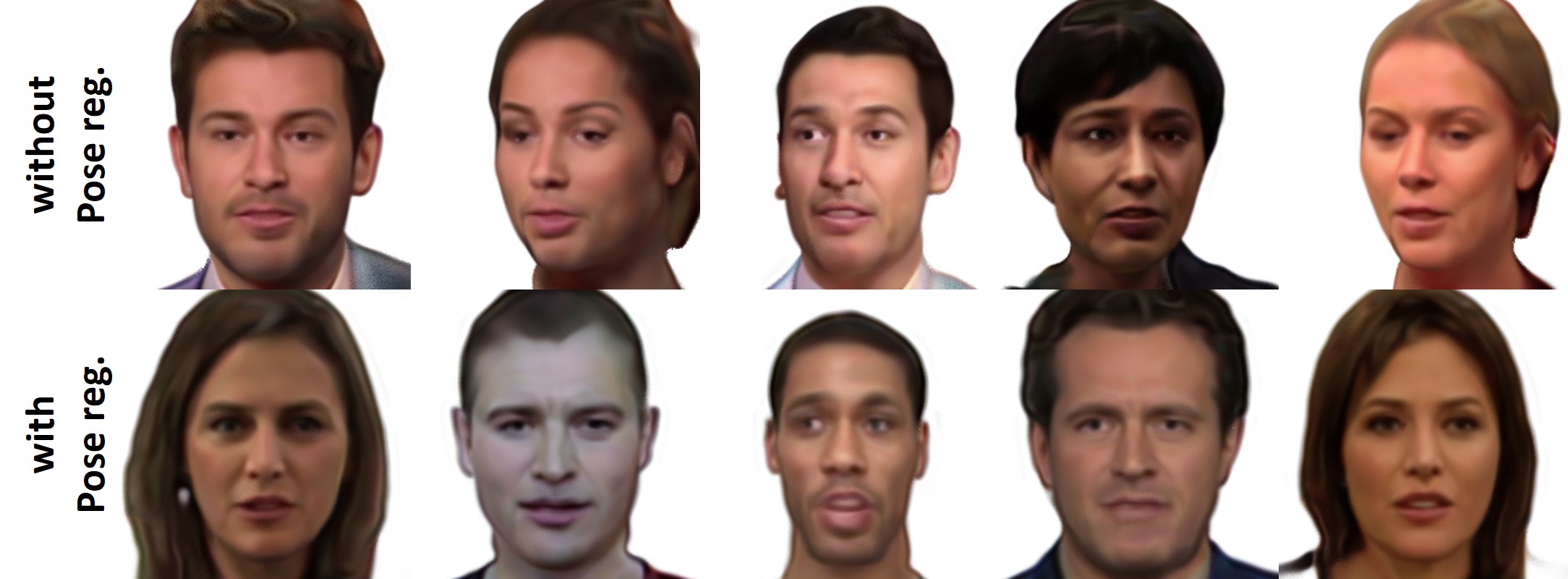}
\vspace{-0.6cm}
\caption{Ablative analysis of the pose regularization loss on VoxCeleb2.
All images are rendered with a fixed frontal camera. 
Without this loss, the head pose changes even though the camera is fixed. 
Pose regularization loss helps in better disentanglement of the 3D scene from the camera viewpoint.}
\label{fig:ablative_pose}
\vspace{-0.4cm}
\end{figure}

We evaluate the inverse deformation network by visualizing the dense correspondences in Fig.~\ref{fig:correspondense}. 
We first provide image-level annotations on one image generated by D3D. 
These annotations can then be transferred to any other sample of the model using the dense correspondences. 
Our model learns correspondences without any explicit supervision, even for objects with large deformations. 
This enables applications such as one-shot segmentation transfer and keypoint annotation. 
In Fig.~\ref{fig:ablative_pose}, we further visualize the effectiveness of the proposed pose regularization loss.
Without this loss, the geometry component tends to entangle the geometry with camera viewpoint.
This is most evident when training with VoxCeleb2~\cite{Chung18b} dataset. 
While this dataset has larger pose varrations compared to FFHQ~\cite{Karras_2019_CVPR}, we used the same prior pose distribution, which could lead to the geometry network also compensating for the inaccurate distribution. 
Our loss term disambiguates pose and the 3D scene, reducing the burden of estimating a very accurate pose distribution. 

\begin{figure}[t!]
\centering
\includegraphics[width=\linewidth]{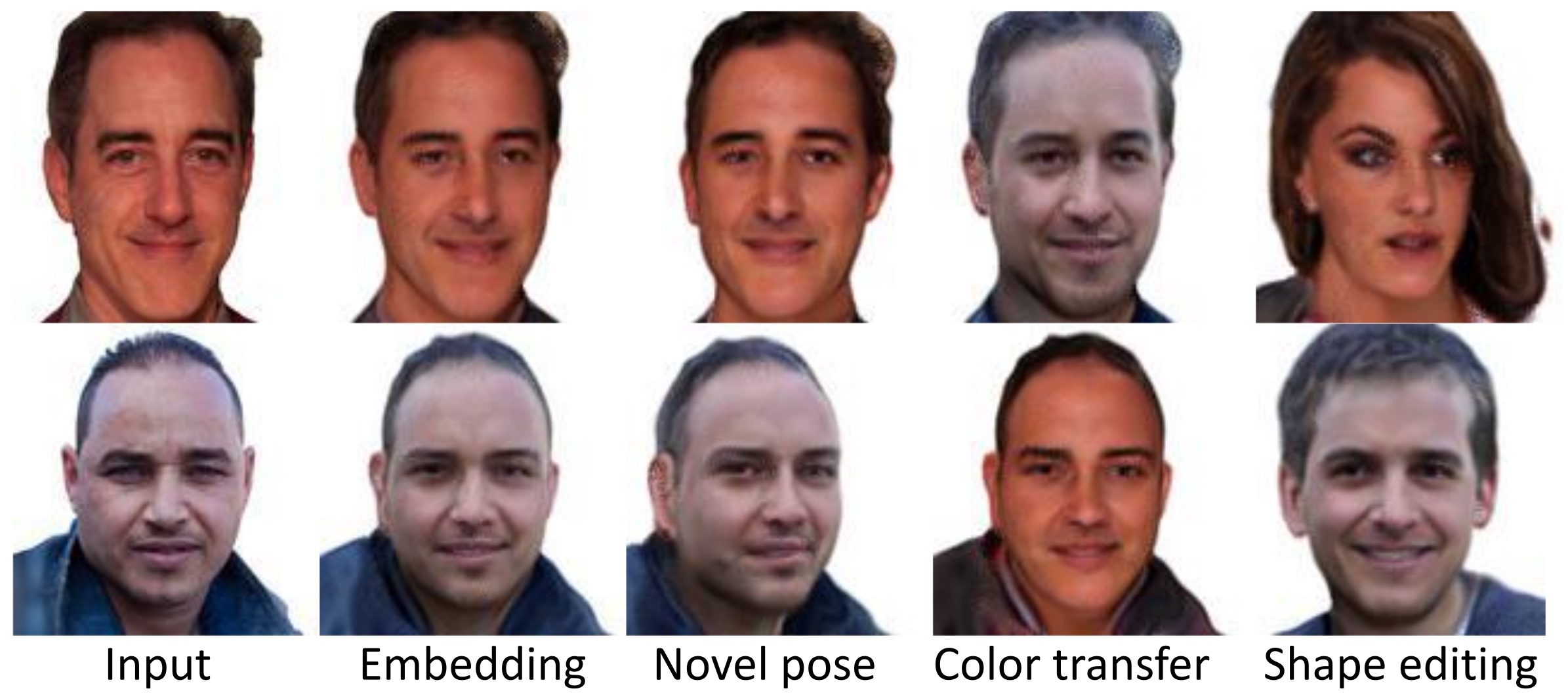}
\vspace{-0.6cm}
\caption{ Given real images (col 1), we can embed them in our GAN space (col 2).  This enables novel view synthesis (col 3),  color transfer from the other real image (col 4), or shape editing using  a random sample from the GAN.}
\vspace{-0.4cm}
\label{fig:embedding}
\end{figure}

We also show embeddings of real images~\cite{Shih14} in Fig.~\ref{fig:embedding}. %
Using our inversion method, we can achieve high-quality embeddings which enables several applications such as pose editing, shape editing, and appearance editing. 
For example, we can transfer the appearance of one portrait image to another, without changing the geometry. We recommend readers refer to the supplementary material for more  results. %
\vspace{-0.3cm}
\paragraph{Quantitative Results}
\begin{table}[]
\centering
\begin{tabular}{lcccc}
\toprule
     & FFHQ & VoxCeleb2 & Cats & Carla \\
\midrule
GRAF~\cite{Schwarz2020NEURIPS} & 43.32     &    35.28      & 22.64     & 37.53       \\
Ours & \textbf{28.18}     &    \textbf{16.51}      &  \textbf{16.96}    &  \textbf{31.13}    \\
\bottomrule
\end{tabular}
\caption{Quantitative comparisons using the FID score metric (a lower value is better). We outperform GRAF on all datasets. }
\label{tab:fid}
\vspace{-0.2cm}
\end{table}
\begin{table}[]
\centering
\begin{tabular}{lcccc}
\toprule
     & \multicolumn{1}{c}{\begin{tabular}[c]{@{}c@{}}$\pi$-GAN \\ \scalebox{0.8}{\cite{chanmonteiro2020pi-GAN}} \end{tabular}} & \multicolumn{1}{c}{\begin{tabular}[c]{@{}c@{}}Ours \\ \scalebox{0.8}{(256-dim)}\end{tabular}} & \multicolumn{1}{c}{\begin{tabular}[c]{@{}c@{}}Ours \\ \scalebox{0.8}{(No inverse)}\end{tabular}} & \multicolumn{1}{c}{\begin{tabular}[c]{@{}c@{}}Ours \\ \scalebox{0.8}{(Complete)}\end{tabular}}  \\ 
\midrule
FFHQ &    \multicolumn{1}{c}{\textbf{13.22}}    &   \multicolumn{1}{c}{13.98} &  \multicolumn{1}{c}{19.99}   & \multicolumn{1}{c}{28.18}     \\
\bottomrule
\end{tabular}
\vspace{-0.2cm}
\caption{Ablation results on FFHQ~\cite{Karras_2019_CVPR} with different baselines, using FID scores. 
Our complete method enables disentanglement of geometry from appearance, in addition to enabling dense correspondences. 
This leads to a loss of quality, as seen here. }
\label{tab:quant_ablative}
\end{table}

\begin{table}[]
\centering
\begin{tabular}{ll|ll}
\toprule
      & \begin{tabular}[c]{@{}l@{}}\scalebox{0.9}{Appearance}\\ \scalebox{0.9}{Consistency} $\downarrow$\end{tabular} & \begin{tabular}[c]{@{}l@{}}\scalebox{0.9}{Geometry} \\ \scalebox{0.9}{Consistency} $\downarrow$ \end{tabular} & \begin{tabular}[c]{@{}l@{}}\scalebox{0.9}{Appearance}\\ \scalebox{0.9}{Variation} $\uparrow$\end{tabular} \\ \hline
$\pi$-GAN & 0.15                                                                                  & 0.96                                                                                & 0.15                                                                               \\
GRAF   & 0.17                                                                                  & \textbf{0.08}                                                                       & 0.04                                                                               \\
Ours \scalebox{0.8}{(256-dim)} 
 & 0.13 & 0.11 & 0.07 \\
Ours \scalebox{0.8}{(No inverse)} 
 & 0.06 & 0.40 & 0.15 \\
Ours \scalebox{0.8}{(Complete)}  & \textbf{0.05}                                                                         & 0.39                                                                                & \textbf{0.16}                                                                      \\ \hline
\end{tabular}
\vspace{-0.2cm}
\caption{
Evaluation of disentanglement.
The first column measures appearance consistency for images rendered with the same appearance code and different geometry codes. 
The second column measures the geometry consistency for images rendered with the same geometry code and different appearance codes. 
The third column measures the appearance variation for such images, higher implies more variation captured in the model. }
\label{tab:disentangle}
\vspace{-0.2cm}
\end{table}

We first provide the commonly reported FID scores~\cite{heusel2017gans} for images generated by our model, as well as those for GRAF~\cite{Schwarz2020NEURIPS} in Table~\ref{tab:fid}. 
The FID scores are computed using $8k$ image samples. 
Our approach outperforms GRAF on all datasets. 
We also perform an ablation study on FFHQ with several baselines in Table~\ref{tab:quant_ablative}. 
``Ours (256-dim)'' is a baseline that implements the design of GRAF in our training framework, i.e., $\mathbf{N}_G(\cdot)$ directly provides a 256-dimensional vector as output, which is sent to $\mathbf{N}_A(\cdot)$ and $\mathbf{N}_C(\cdot)$.
Other network architecture and training details are equivalent to our method. 
However, this design makes it infeasible to use the pose consistency loss and inverse deformations, so we disable them.
This framework achieves a lower FID compared to our complete model, %
however, it does not achieve high-quality disentanglement due to the same reasons as for GRAF, see the supplemental document.
``Ours (No inverse)'' is our method without the inverse deformations. 
This architecture constraints the network by limiting $\mathbf{N}_G(\cdot)$ to output a 3-dimensional deformation of coordinates.
This leads to good disentanglement at the cost of slightly higher FID. 
``Ours (Complete)'' further incorporates the inverse deformation network, which allows us to compute dense correspondences. 
While this enables broader interesting applications, it again comes at a cost of higher FID scores due to stronger regularization of the deformation field.
We also report the FID score of $\pi$-GAN~\cite{chanmonteiro2020pi-GAN}, which is comparable to our 256-dimensional baseline. 
Note that $\pi$-GAN does not enable any disentanglement between the geometry and appearance components.

We quantitatively evaluate the quality of disentanglement in Table~\ref{tab:disentangle}. 
We describe two novel metrics to evaluate this.
To evaluate the consistency of appearance with changing geometry, we measure the standard deviation of the average color in a semantically well-defined region, which could be obtained via an off-the-shelf segmentation model~\cite{yu2018bisenet}. 
We use the hair region for human heads to compute this metric for networks trained on FFHQ~\cite{Karras_2019_CVPR}. 
We sample $100$ images from the GAN with a fixed appearance code and varying geometry codes. 
The standard deviation of the average hair color can be used as a metric, as a lower value would imply consistent appearance across different shapes. 
We compute this standard deviation for $10$ appearance codes and report the average over the $10$ values. 
Our approach significantly outperforms GRAF~\cite{Schwarz2020NEURIPS} and $\pi$-GAN~\cite{chanmonteiro2020pi-GAN}. 
Since $\pi$-GAN does not have different appearance and geometry codes, we simply sample $1000$ images from their model and use the numbers as a baseline.

To evaluate the geometry consistency for a fixed geometry code with varying appearances, we use sparse facial keypoints for evaluation. 
We measure the standard deviation of $66$ facial landmarks computed using an off-the-shelf tool~\cite{saragih2011deformable} across $100$ samples with a shared geometry code and different randomly sampled appearance codes. 
We render all images in the same pose, in order to eliminate additional factors of variance. 
This evaluation is repeated for $10$ different geometry codes and the error is averaged over these geometry codes, and over the $66$ landmarks. 
A lower number with the geometry consistency metric implies that varying the appearance code is less likely to cause geometry change in the image.
While we outperform the $\pi$-GAN baseline, GRAF~\cite{Schwarz2020NEURIPS} achieves a better score. 
This is due to the fact that the appearance variations are limited for GRAF, as the appearance information also leaks into the geometry component. 
We further evaluate this using an appearance variation metric for these images. 
This metric is defined exactly the same as the appearance consistency metric. 
Specifically, for the set of images, we calculate the  standard deviation over the average hair color over the 100 images with different appearance codes, and average over the 10 geometry codes.
As shown in Table~\ref{tab:disentangle}, our method achieves the highest value, implying that our appearance component better captures the appearance variations of the dataset. 
{We also evaluate both baselines using these metrics. As expected, the ``256-dim`` baseline performs similar to GRAF, while the numbers are similar without the inverse network}

\section{Conclusion \& Discussion}
We have presented an approach to learn disentangled 3D GANs from monocular images.
In addition to disentanglement, our formulation enables the computation of dense correspondences, enabling exciting applications. 
Although we have demonstrated compelling results, our method has several limitations. 
Like other 3D GANs, our results do not reach the photorealism quality and image resolutions of 2D GANs. 
The disentanglement and correspondences come at the cost of a  drop in image quality (see Table~\ref{tab:quant_ablative}).
In addition, we use an off-the-shelf background segmentation tool which limits us from being completely unsupervised. 
Nevertheless, our approach achieves high image quality and disentanglement, significantly outperforming the state of the art. 
We hope that it inspires further work on self-supervised learning of 3D generative models. 

\let\thefootnote\relax\footnotetext{
	\textbf{Acknowledgements:}
This work was partially supported by the ERC Consolidator Grant 4DReply (770784), the Brown Institute for Media Innovation, and the Israel Science Foundation (grant No. 1574/21).
}

{\small
\bibliographystyle{ieee_fullname}
\bibliography{egbib}
}
\clearpage

\appendix

\section{Training Details}
\paragraph{Network Architecture}
Our generator network consists of a geometry deformation network $\mathbf{N}_G$,  an appearance network $\mathbf{N}_A$, and a canonical geometry network $\mathbf{N}_C$.
Both $\mathbf{N}_G$ and $\mathbf{N}_A$ include a mapping network and a main network following the design of $\pi$-GAN~\cite{chanmonteiro2020pi-GAN}. 
The mapping networks are implemented as  MLPs with {LeakyReLU} activations, see Table~\ref{tab:mapping}. 
The randomly sampled inputs $\rvz_G \in \mathbb{R}^{256}$ and $\rvz_A \in \mathbb{R}^{256}$ are used as inputs to the mapping networks. 
The output of the mapping networks are one-dimensional vectors of dimensions $256 \times 2 \times d_G$ and $256 \times 2 \times d_A$, where $d_G$ and $d_A$ are the number of {SIREN} layers in the main networks of $\mathbf{N}_G$ and $\mathbf{N}_A$ respectively. 
The main networks are implemented as MLPs with {SIREN} layers~\cite{sitzmann2019siren} and {FiLM} conditioning~\cite{perez2018film}, see Table~\ref{tab:geometry} and Table~\ref{tab:appearance}. 
Each layer of the main network receives one $256 \times 2$-dimensional component of the output of the mapping network. 
The canonical network $\mathbf{N}_C$ does not receive any input other than the co-ordinates in the canonical space.
We follow the initialization method of \cite{sitzmann2019siren} for $\mathbf{N}_G$, $\mathbf{N}_A$, and $\mathbf{N}_C$, where the first layer is initialized with larger values. 
The final layer of $\mathbf{N}_G$ is initialized such that the deformations at the first iteration are all zeros. 
The inverse deformation network is implemented exactly as $\mathbf{N}_G$, except that it receives the input in the canonical space and models the inverse deformation. %
As for the discriminator, we adopt the same model architecture as in~\cite{chanmonteiro2020pi-GAN}, which is a convolutional neural network with residual connections~\cite{resnet_16} and CoordConv layers~\cite{liu2018intriguing}.

As explained in the main paper, we control the level of disentanglement using the number of {SIREN} layers in $\mathbf{N}_G$  and $\mathbf{N}_A$, i.e., $d_G$ and $d_A$, respectively. 
We set $d_G=5$ and $d_A=3$ for FFHQ~\cite{Karras_2019_CVPR}, VoxCeleb2~\cite{Chung18b}, and Cats~\cite{zhang2008cat}. 
For Carla~\cite{dosovitskiy2017carla}, we set $d_G=3$ and $d_A=6$.
We will show results where changing the relative depths of these networks can lead to poor disentanglement.

\begin{table}[]
\centering
\resizebox{8.5cm}{1.05cm}{
\begin{tabular}{llll}
Input & Layer & Activation & Output Dim. \\ \hline
$\rvz_G$ or $\rvz_A$  & Linear & LeakyReLU (0.2) & 256 \\
- & Linear & LeakyReLU (0.2) & 256 \\
- & Linear & LeakyReLU (0.2) & 256 \\
- & Linear & None & 256 $\times$ 2 $\times (d_G \text{~or~} d_A)$
\end{tabular}
}
\caption{Mapping Network, denoted as {Map}($\cdot$). We use separate mapping networks for the geometry and appearance networks. }
\label{tab:mapping}
\end{table}

\begin{table}[]
\centering
\begin{tabular}{llll}
Input  & Layer    & Activation & Output Dim. \\ \hline
$\mathbf{x'}$ & Linear & Sine     & 256         \\
- & Linear & Sine     & 256         \\
- & Linear & Sine     & 256         \\
- & Linear & Sine     & 256         \\
- & Linear & None       & 1          
\end{tabular}
\caption{Canonical Network, denoted as $\mathbf{N}_C$($\cdot$). The input $\mathbf{x'}$ is a point in the canonical space, computed using the goemetry deformation network. }
\label{tab:canonical}
\end{table}

\begin{table}[]
\centering
\begin{tabular}{llll}
Input                                                                                              & Layer    & Activation & Output Dim. \\ \hline
$\mathbf{x}$,  {Map}($\rvz_G$) & Linear & FiLM+Sine     & 256         \\
                                                                                          -,  {Map}($\rvz_G$)         & ...      & ...        & ...         \\
                                                                                          -,  {Map}($\rvz_G$)         & ...      & ...        & ...         \\
                                                                                          -,  {Map}($\rvz_G$)         & Linear & None       & 3          
\end{tabular}
\caption{Geometry Deformation Network, denoted as $\mathbf{N}_G$($\cdot$). The input $\mathbf{x}$ is a point in the deformed or world space. The output can be added to $\mathbf{x}$ to compute $\mathbf{x'}$, the corresponding 3D point in the canonical space.
The output of the shape mapping network is also provided as input for each layer. }
\label{tab:geometry}
\end{table}

\begin{table}[]
\centering
\begin{tabular}{llll}
Input                                                                                              & Layer    & Activation & Output Dim. \\ \hline
$\mathbf{x'}$,  {Map}($\rvz_A$) & Linear & FiLM+Sine     & 256         \\
                                                                                          -,  {Map}($\rvz_A$)         & ...      & ...        & ...         \\
                                                                                          -,  {Map}($\rvz_A$)         & ...      & ...        & ...         \\
                                             
                                             -,  {Map}($\rvz_A$), $\mathbf{d}$         & Linear      & FiLM+Sine       & 256         \\
                                                                                          -,  {Map}($\rvz_A$)         & Linear & Sigmoid       & 3          
\end{tabular}
\caption{Appearance Network, denoted as $\mathbf{N}_A$($\cdot$). The input $\mathbf{x'}$ is a point in the canonical space, computed using the goemetry deformation network. The output is the color at this point. The other inputs are the output of the color mapping network, and the viewing direction. }
\label{tab:appearance}
\vspace{0.5cm}
\end{table}

\paragraph{Hyperparameters}
\begin{table}[]
\centering
\begin{tabular}{lll}
Hyperparameter & Dataset & Value \\ \hline
$\lambda$ & FFHQ & 1.0 \\
 & VoxCeleb2 & 1.0 \\
 & Cats & 0.5 \\
 & Carla & 10.0 \\ \hline
$\lambda_\text{pose}$ & FFHQ & 50.0 \\
 & VoxCeleb2 & 50.0 \\
 & Cats & 5.0 \\
 & Carla & 50.0 \\ \hline
$\lambda_\text{img}$ & FFHQ & 0.001 \\
 & VoxCeleb2 & 0.001 \\
 & Cats & 0.001 \\
 & Carla & 0.001 \\ \hline
$\lambda_\text{inv}$ & FFHQ & 1.0 \\
 & VoxCeleb2 & 1.0 \\
 & Cats & 1.0 \\
 & Carla & 1.0
\end{tabular}
\caption{Hyperparameters of our method.}
\label{tab:hyperparameters}
\end{table}
\begin{table*}[]
\centering
\begin{tabular}{llllll}
\hline
Dataset   & Iteration (in k) & Batch Size & Image Size & $G_{lr}$ & $D_{lr}$ \\ \hline
FFHQ      & 0-20             & 208        & 32         & 2e-5     & 2e-4     \\
          & 20-60            & 52         & 64         & 2e-5     & 2e-4     \\
          & 60-            & 52         & 64         & 1e-5     & 1e-4     \\ \hline
VoxCeleb2 & 0-20             & 208        & 32         & 2e-5     & 2e-4     \\
          & 20-60            & 52         & 64         & 2e-5     & 2e-4     \\
          & 60-            & 52         & 64         & 1e-5     & 1e-4     \\ \hline
Cats      & 0-10             & 208        & 32         & 6e-5     & 2e-4     \\
          & 10-            & 52         & 64         & 6e-5     & 2e-4     \\ \hline
Carla     & 0-10             & 60         & 32         & 4e-5     & 4e-4     \\
          & 10-26            & 20         & 64         & 2e-5     & 2e-4     \\
          & 26-            & 18         & 128        & 10e-6    & 10e-5   \\ \hline
\end{tabular}
\caption{Training curriculum}
\label{tab:curriculum}
\end{table*}

We describe the hyperparamters used in our method in Table~\ref{tab:hyperparameters}.
The training curriculum is described in Table~\ref{tab:curriculum}. Our networks are trained in a coarse-to-fine manner. 

\paragraph{Embedding Architecture}
Our encoder network consists of a pretrained ResNet-18~\cite{resnet_16} as the backbone.
We add two linear layers to regress the camera pose and latent vectors.
Inspired by $\pi$-GAN~\cite{chanmonteiro2020pi-GAN}, we learn to directly regress the frequencies and phase shifts, i.e., the output space of the mapping networks for the geometry and appearance components.
We train the encoder on FFHQ~\cite{Karras_2019_CVPR}.
We set $\lambda_\text{perc} = 1$ and $\lambda_\text{reg} = 10$ and use a learning rate of $0.01$.

\begin{figure}
\centering
\includegraphics[width=\linewidth]{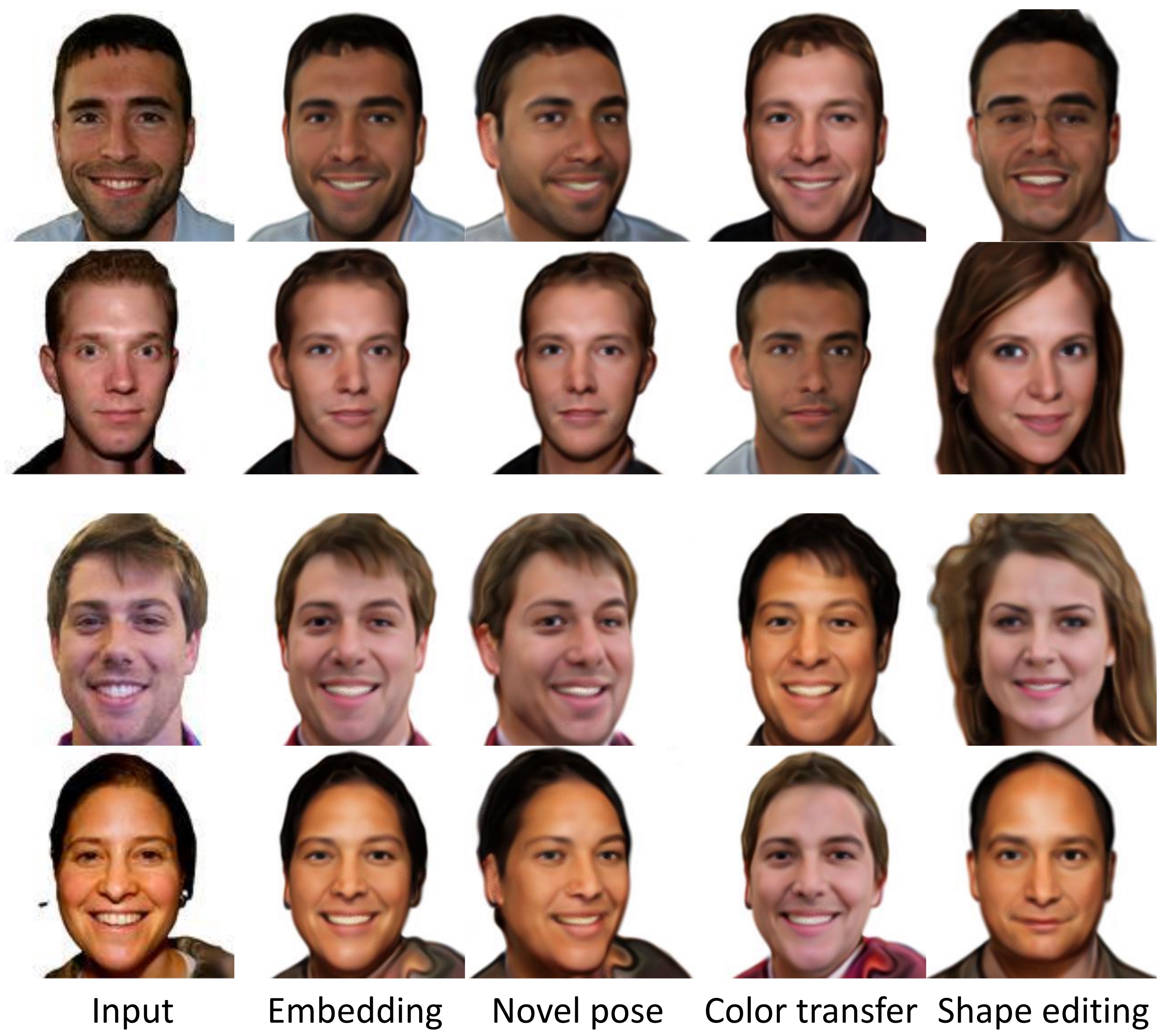}
\caption{Given real images (col 1), we can embed them in our GAN space (col 2).  This enables novel view synthesis (col 3),  color transfer from the other real image (col 4), or shape editing using  a random sample from the GAN. For color transfer results in col 4, we transfer the embedded color between 2 pairs ( rows 1,2 and rows 3,4). }
\label{fig:embedding_appln}
\end{figure}

At test time, to further improve the result, we fine-tune the regressed latent vectors using iterative optimization for $1.8k$ iterations with a learning rate of $0.01$.
We finally fine-tune the generator network for another $200$ iterations with a learning rate of $0.0001$.
We show that this strategy leads to high-quality results without degrading the disentanglement properties (see Fig.~\ref{fig:embedding_appln}) of the generator.

\begin{figure}
\centering
\includegraphics[width=\linewidth]{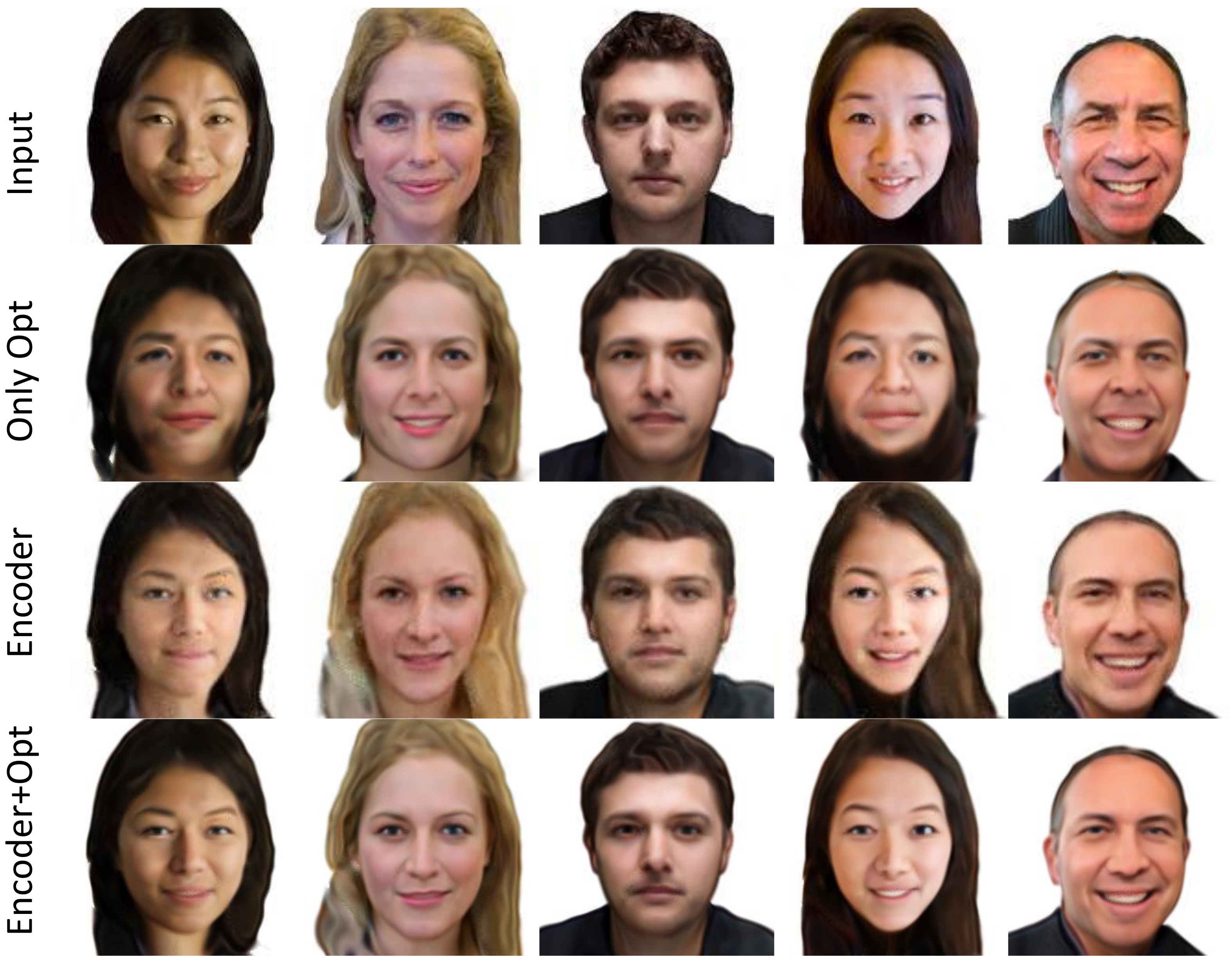}
\caption{Here we show that our embedding method which uses encoder output as initialization (row 3) results in higher-quality output (row 4) compared to optimization-only approach (row 2) for real in-the-wild input images (row 1).}
\label{fig:embedding_abl}
\end{figure}

We also show that this approach works better than optimization-only method (see Fig.~\ref{fig:embedding_abl}), where we iteratively optimize for the latent vectors and camera pose using reconstruction loss. 
For optimization-only approach, we update the latent vectors and camera pose while keeping the GAN fixed for $1.8k$ iterations with a learning rate of $0.01$. And then finetune GAN as well for another $200$ iterations with a learning rate of $0.0001$.
We can observe (Fig.~\ref{fig:embedding_abl},  ~\ref{fig:embedding_appln}) that using encoder initialization helps obtain better results while still preserving the disentanglement properties of our model.

\label{sec:training}

\section{Results}
\label{sec:results}
\begin{table}[t]
\centering
\begin{tabular}{lc}
\toprule
 & \multicolumn{1}{l}{Pose Consistency $\downarrow$} \\ \midrule
pi-GAN & 0.34 \\
\begin{tabular}[c]{@{}l@{}}Ours \scalebox{0.8}{(no pose reg.)}\end{tabular} & 0.16 \\
Ours & \textbf{0.03} \\
\bottomrule
\end{tabular}
\caption{Quantitative evaluation of pose consistency. 
Pose consistency is measured as the standard deviation of the 3D yaw-component of head pose computed over $1000$ images rendered from a fixed camera. 
The pose regularization significantly improves pose consistency, helping disentangle the camera pose from the scene. }
\label{tab:pose}
\end{table}
\paragraph{Qualitative Results}
\begin{figure*}
\includegraphics[width=\textwidth]{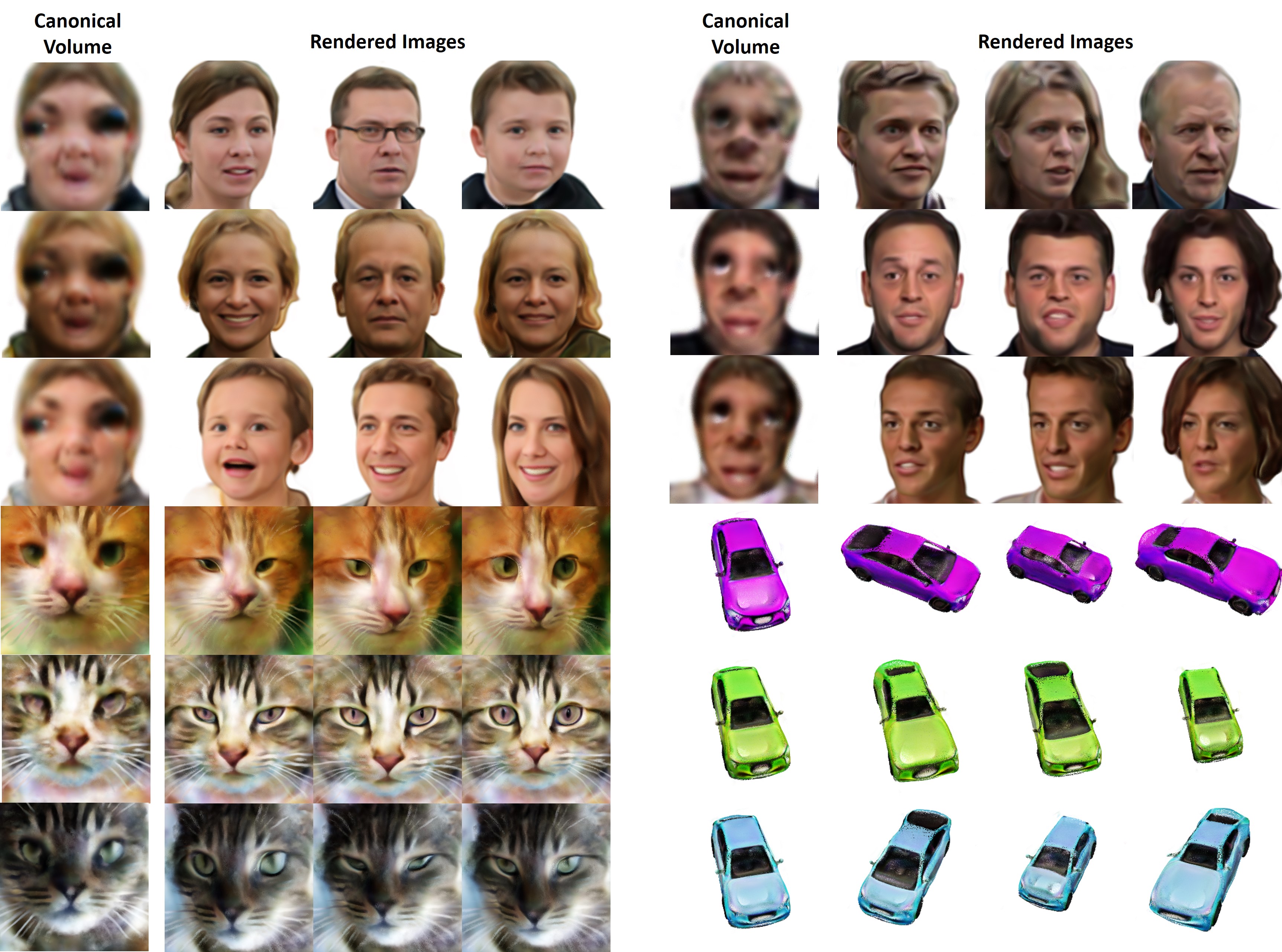}
\caption{Results of our method on FFHQ (top-left), VoxCeleb2 (top-right), Cats (bottom-left) and Carla (bottom-right). Each row shows the canonical volume, and multiple rendered images with the same appearance and pose, but with different geometry. All canonical volumes for a dataset are rendered from the same pose. Notice that only the color of the canonical volume changes.  }
\vspace{0.7cm}
\label{fig:canonical}
\end{figure*}
We show more results of our method along with visualizations of the learned canonical volume in Fig.~\ref{fig:canonical}. 
\begin{figure*}
\includegraphics[width=\textwidth]{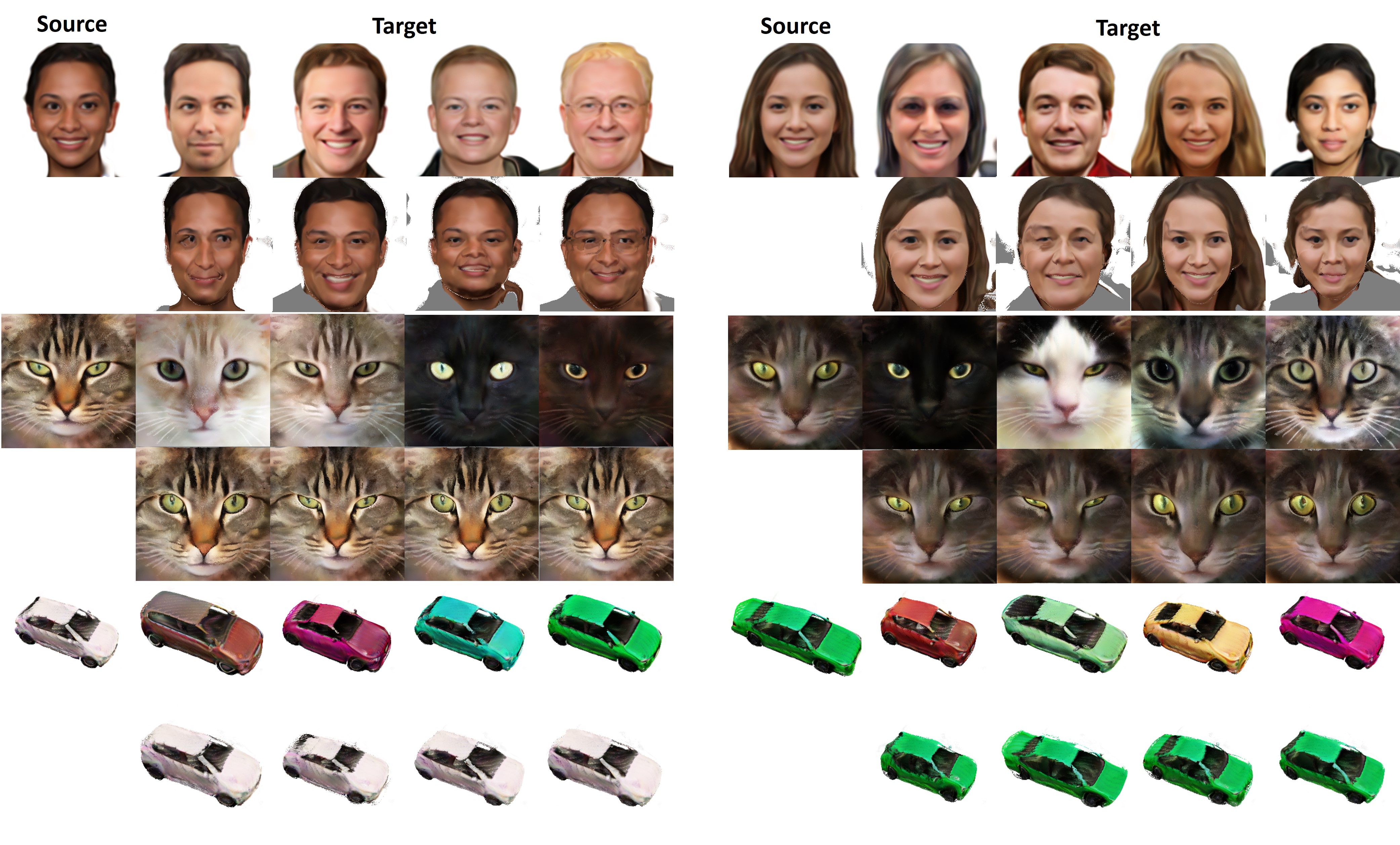}
\caption{Appearance transfer using the learned correspondences. 
For each object class, the first row shows different random samples from our GAN. 
The left-most sample is used as the source texture. 
This texture is transferred to all other samples, visualized in the second row. 
Note that we only the source image, and not the full 3D model, in order to visualize pixel-to-pixel correspondences. 
We can faithfully transfer the source appearance while preserving the target geometry. 
Also note that not all pixels in the target image have a valid correspondence to the source image. 
For example, if the shirt is not visible in the source image, the shirt pixels in the target image do not have a valid correspondence. 
Thus, only the pixels whose corresponding points are visible in the source image achieve the correct appearance transfer. 
This visualization shows the applicability of our approach to various applications, such as one-shot semantic segmentation and sparse keypoint detection. 
}
\vspace{0.5cm}
\label{fig:appearance_transfer}
\end{figure*}
We present more visualizations of the learned correspondences in Fig.~\ref{fig:appearance_transfer}. %
The appearance of one sample is transferred to another using the correspondences. 
This shows the applicability of the correspondences for any task where one image annotation can be transferred to all other samples of the model. 
\begin{figure}
    \centering
    \includegraphics[width=0.48\textwidth]{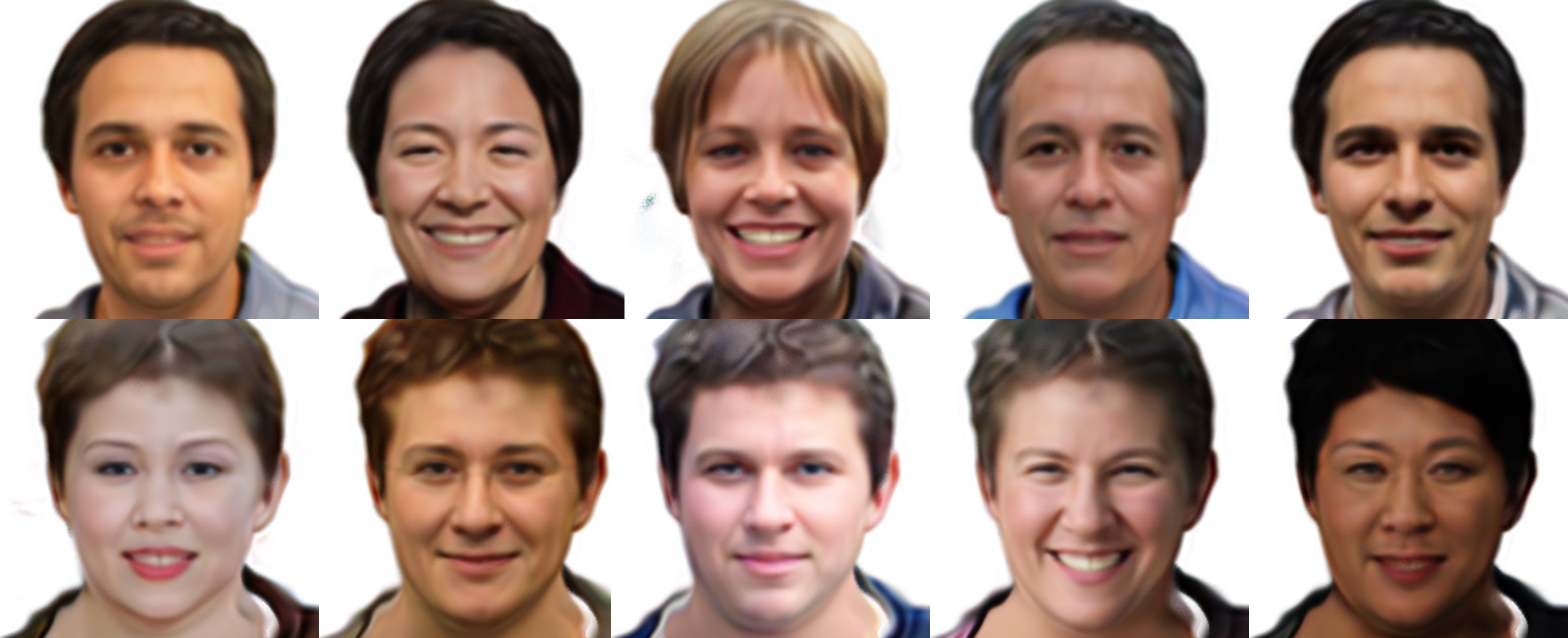}
    \caption{Results on FFHQ with a larger appearance network. Each row shows results with a fixed geometry and different appearances. 
    With a large appearance network, geometric features such as expressions can be compensated incorrectly by the appearance component. }
    \vspace{0.5cm}
    \label{fig:large_color}
\end{figure}
As mentioned earlier, the level of disentanglement is controlled using the relative depths of the geometry and appearance networks. 
We show in Fig.~\ref{fig:large_color} that  a large appearance network can lead to lower-quality disentanglement, where geometric features such as expressions are compensated by the appearance component. 
We set $d_G = 3$ and $d_A = 5$ for these results. 
\begin{figure}
    \centering
    \includegraphics[width=0.48\textwidth]{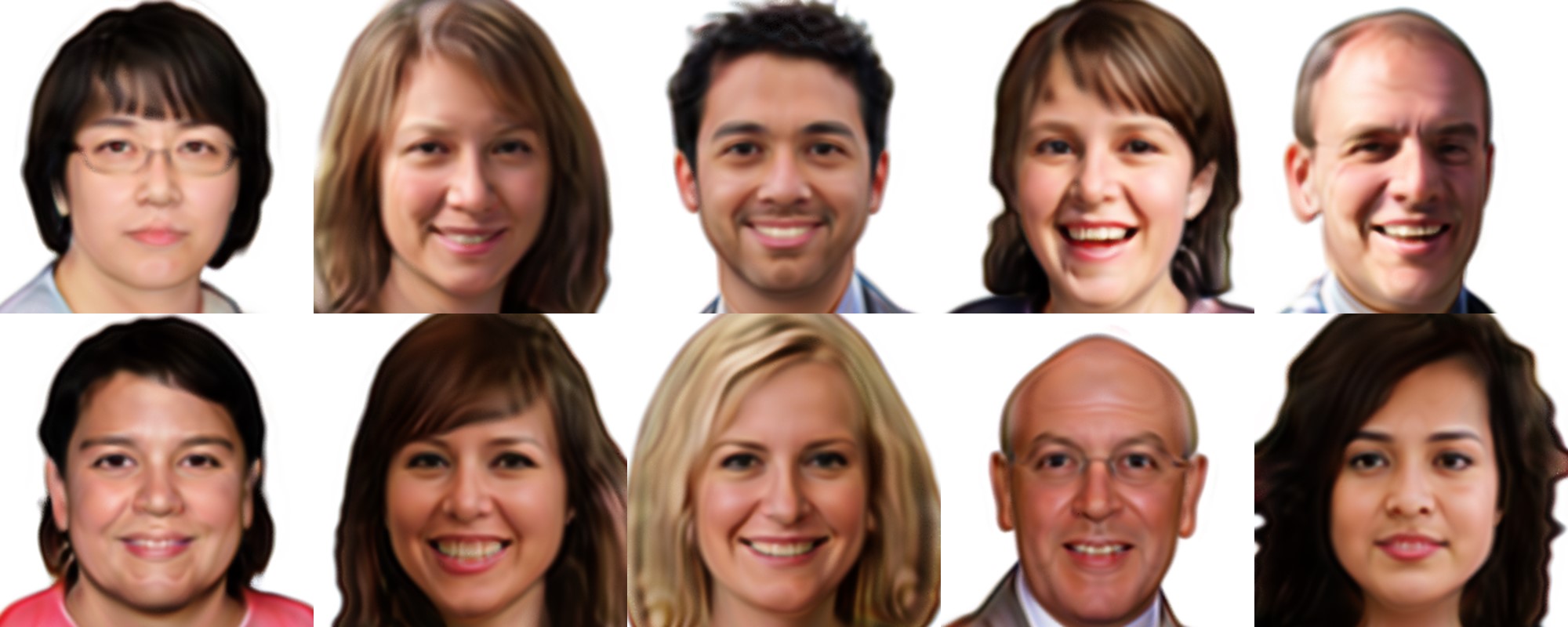}
    \caption{Results of the 256-baseline on FFHQ. Each row shows results with a fixed appearance and different geometry. 
    This baseline uses a 256-dimension vector as input to the canonical volume. 
    This results in poor disentanglement, where changing the geometry also changes the appearance. %
    GRAF~\cite{Schwarz2020NEURIPS} uses a similar design choice and thus, suffers from the same limitation. 
    }
    \vspace{0.5cm}
    \label{fig:256baseline}
\end{figure}
In the main paper, we presented quantitative results of a baseline where the canonical network receives a high-dimensional input like GRAF~\cite{Schwarz2020NEURIPS}. 
Fig.~\ref{fig:256baseline} shows qualitative results of this baseline. 
As explained in the main paper, this baseline has similar limitations as GRAF, where the geometry network also changes the appearance of the object. 
\begin{figure}
    \centering
    \includegraphics[width=0.45\textwidth]{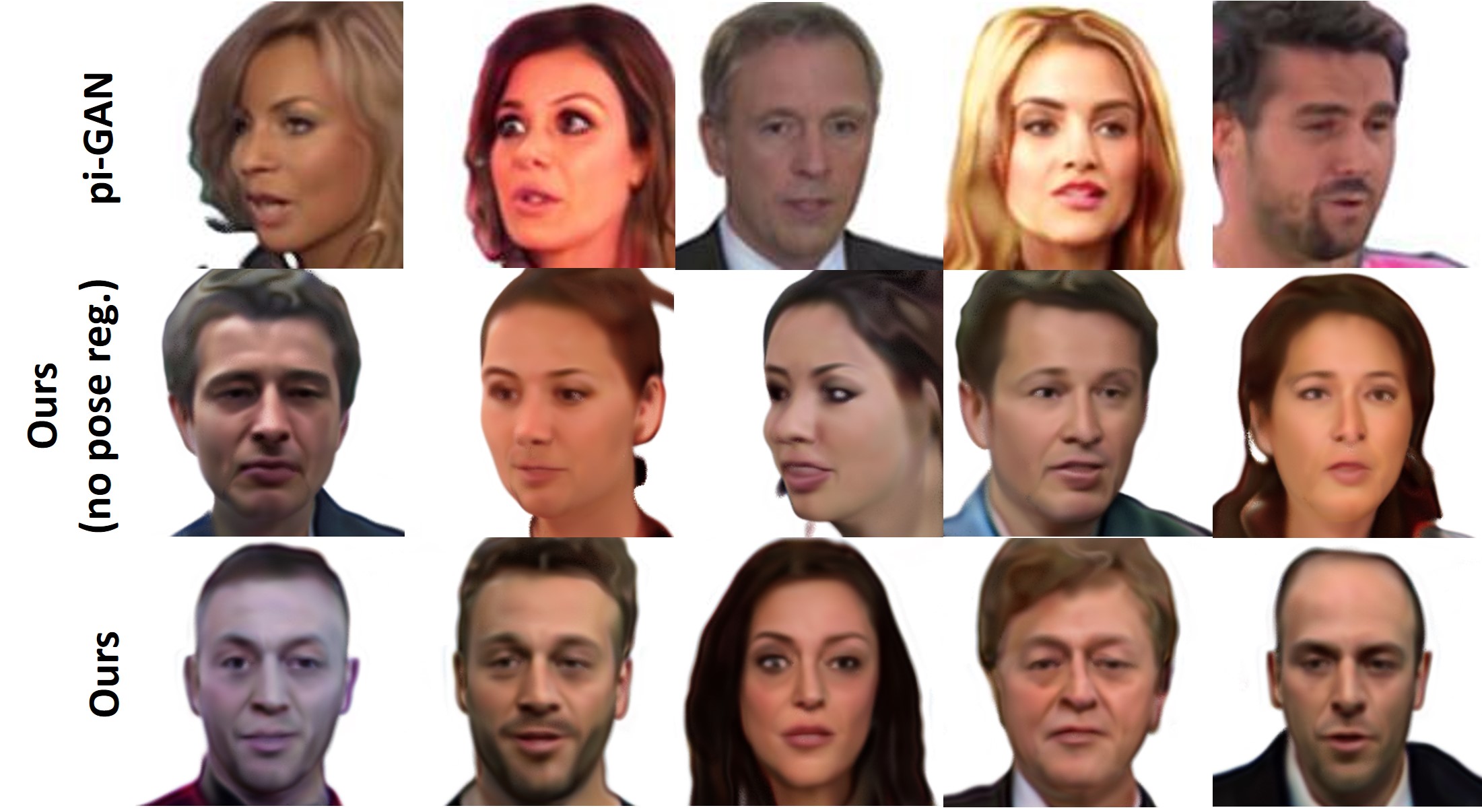}
    \caption{Evaluation of our pose regularization loss on VoxCeleb2. 
    All images are rendered with a fixed frontal camera. 
    Without pose regularization, the model cannot disentangle between the scene and the camera pose.
    This issue is also evident in pi-GAN. 
    }
    \label{fig:pose}
\end{figure}
Fig.~\ref{fig:pose} shows more results for evaluation of the pose regularization. 
Without our proposed regularization, the model does not properly disentangle the object and the camera pose. 
This limitation is also shared with $\pi$-GAN~\cite{chanmonteiro2020pi-GAN}.
\begin{figure}
    \centering
    \includegraphics[width=\linewidth]{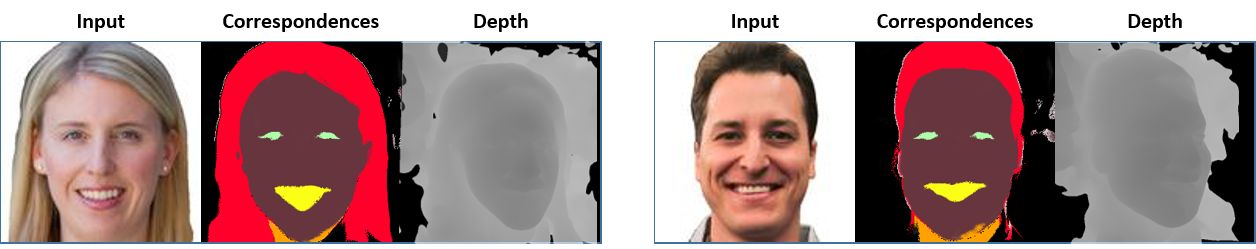}
    \caption{\AT{Results on real images. Reference from Fig.5-main is used for correspondences. Depth is rendered from a novel view.}}
    \label{fig:real_corr}
\end{figure}
\AT{We further show some results of correspondence and depth visualizations on real images in Fig.~\ref{fig:real_corr}.
Unlike the encoders used in other results, we trained the encoder for this result on the generator which was trained with the inverse network.}
\begin{figure}
    \centering
    \includegraphics[width=\linewidth]{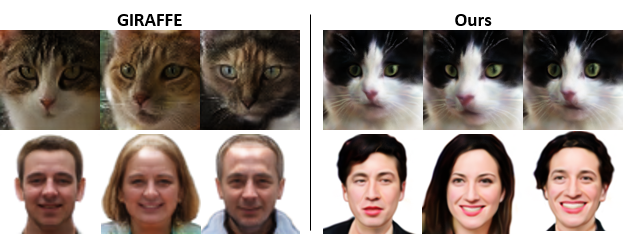}
    \caption{\AT{Comparisons with GIRAFFE. Visualized are three images with the same appearance code but different geometry codes.}}
    \label{fig:giraffe}
\end{figure}
\AT{We also compare to GIRAFFE~\cite{niemeyer2021giraffe} in Fig.~\ref{fig:giraffe}. 
Our method maintains the consistency of both pose and shape components better. 
Quantitatively, GIRAFFE achieves similar scores compared to our method on FFHQ using the metrics defined in the main paper. 
It achieves an appearance consistency score of 0.05, geometry consistency score of 0.32, and appearance variation score of 0.09.
However, ours results have better multi-view consistency, and better qualitative disentanglement as shown in Fig.~\ref{fig:giraffe}. 
} 
\begin{figure*}
\centering
\includegraphics[width=0.9\textwidth]{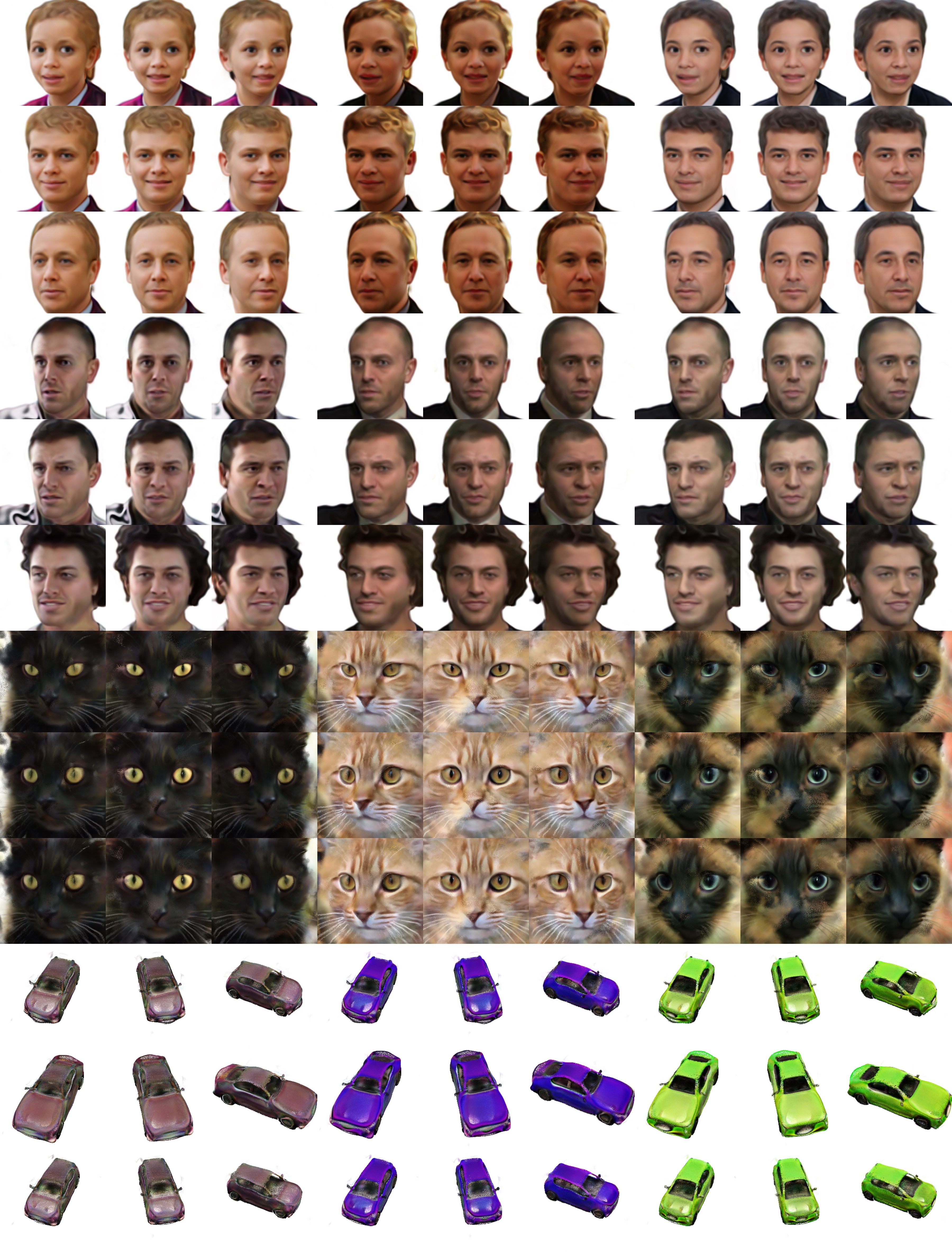}
\caption{More results of our method on FFHQ (rows 1-3), VoxCeleb2 (rows 4-6), Cats (rows 6-8) and Carla (rows 10-12). Each row shows a fixed geometry with three different appearances and poses. }
\label{fig:uncurated}
\end{figure*}
We show several more results of our GAN in Fig.~\ref{fig:uncurated}.

\paragraph{Quantitative results}
\begin{table}[]
\centering
\begin{tabular}{lcccc}
\toprule
     & FFHQ & VoxCeleb2 & Cats \\
\midrule
GRAF~\cite{Schwarz2020NEURIPS} & 25.36     &    21.76      & 18.26       \\
Ours & \textbf{15.87}     &    \textbf{8.86}      &  \textbf{12.35}    \\
\bottomrule
\end{tabular}
\caption{Quantitative comparisons using the FID score metric (a lower value is better) at $64\times64$ image resolution. We outperform GRAF on all datasets. }
\label{tab:fid}
\end{table}
We present FID scores for FFHQ~\cite{Karras_2019_CVPR}, VoxCeleb2~\cite{Chung18b}, and Cats~\cite{zhang2008cat} evaluated at $64\times64$ image resolution in Table~\ref{tab:fid}. All FID scores are calculated using $8$k samples. 
We also present a quantitative evaluation of the pose regularization loss in Table~\ref{tab:pose}.
Specifically, we first render $1000$ images from each method with a fixed camera. 
We then compute the head pose in the rendered results using the Model-based Face Autoencoder (MoFA)~\cite{tewari2017mofa} method. 
The pose consistency metric is computed as the standard deviation over the yaw angles. 
A lower number indicates good disentanglement of the camera pose and the 3D object.
We can see that the proposed pose regularization loss significantly improves such disentanglement.

\end{document}